\documentclass[sigconf, nonacm]{acmart}

\newcommand\vldbdoi{XX.XX/XXX.XX}
\newcommand\vldbpages{XXX-XXX}
\newcommand\vldbvolume{14}
\newcommand\vldbissue{1}
\newcommand\vldbyear{2020}
\newcommand\vldbauthors{\authors}
\newcommand\vldbtitle{\shorttitle} 
\newcommand\vldbavailabilityurl{URL_TO_YOUR_ARTIFACTS}
\newcommand\vldbpagestyle{plain} 
 
\usepackage{algorithmic}
\usepackage{graphicx}
\usepackage{xcolor}
\usepackage{algorithm }
\usepackage{amsmath}
\usepackage{bm}
\usepackage{multirow}
\usepackage{threeparttable}
\usepackage{pifont}
\usepackage[titletoc]{appendix}
\usepackage{enumitem}
\usepackage{cellspace}
\usepackage{subfigure}
\usepackage{color, soul}
\usepackage{xspace}
\usepackage{makecell}
\usepackage{setspace}

\newcommand\projectname{GraphTheta\xspace}
\newcommand\interalname{GeaLearning\xspace}

\begin{document}
\title{\projectname: A Distributed Graph Neural Network Learning System With Flexible Training Strategy}


\author{Yongchao Liu$^{1,*}$, Houyi Li$^{2,\dagger,*}$, Guowei Zhang$^1$, Xintan Zeng$^1$, Yongyong Li$^1$, Bin Huang$^1$, Peng Zhang$^{3,\dagger}$, Zhao Li$^{4,\dagger}$, Xiaowei Zhu$^1$, Changhua He$^1$, Wenguang Chen$^{1,5}$}
\affiliation{%
  \institution{$^1$Ant Group, $^2$Fudan University, $^3$Guangzhou University, $^4$ Zhejiang University, $^5$ Tsinghua University}
  \country{China}
}
\thanks{$^*$ Joint first authors; $^\dagger$ Contributed to this work when the authors worked in Ant Group or Alibaba Group; Correspond to \{yongchao.ly, yuanben.cwg\}@antgroup.com.}

\begin{abstract}
Graph neural networks (GNNs) have been demonstrated as a powerful tool for analyzing non-Euclidean graph data. 
However, the lack of efficient distributed graph learning systems severely hinders applications of GNNs, especially when graphs are big and GNNs are relatively deep.
Herein, we present \projectname, the first distributed and scalable graph learning system built upon vertex-centric distributed graph processing with neural network operators implemented as user-defined functions. This system supports multiple training strategies and enables efficient and scalable big-graph learning on distributed (virtual) machines with low memory.
To facilitate graph convolutions, \projectname puts forward a new graph learning abstraction named NN-TGAR to bridge the gap between graph processing and graph deep learning.
A distributed graph engine is proposed to conduct the stochastic gradient descent optimization with a hybrid-parallel execution, and a new cluster-batched training strategy is supported.
We evaluate \projectname using several datasets with network sizes ranging from small-, modest- to large-scale. 
Experimental results show that \projectname can scale well to 1,024 workers for training an in-house developed GNN on an industry-scale Alipay dataset of 1.4 billion nodes and 4.1 billion attributed edges, with a cluster of CPU virtual machines (dockers) of small memory each (5$\sim$12GB). Moreover, \projectname can outperform DistDGL by up to $2.02\times$, with better scalability, and GraphLearn by up to $30.56\times$. As for model accuracy, \projectname is capable of learning as good GNNs as existing frameworks. To the best of our knowledge, this work presents the largest edge-attributed GNN learning task in the literature.
\end{abstract}

\maketitle

\pagestyle{\vldbpagestyle}
\begingroup\small\noindent\raggedright\textbf{PVLDB Reference Format:}\\
\vldbauthors. \vldbtitle. PVLDB, \vldbvolume(\vldbissue): \vldbpages, \vldbyear.\\
\href{https://doi.org/\vldbdoi}{doi:\vldbdoi}
\endgroup
\begingroup
\renewcommand\thefootnote{}\footnote{\noindent
This work is licensed under the Creative Commons BY-NC-ND 4.0 International License. Visit \url{https://creativecommons.org/licenses/by-nc-nd/4.0/} to view a copy of this license. For any use beyond those covered by this license, obtain permission by emailing \href{mailto:info@vldb.org}{info@vldb.org}. Copyright is held by the owner/author(s). Publication rights licensed to the VLDB Endowment. \\
\raggedright Proceedings of the VLDB Endowment, Vol. \vldbvolume, No. \vldbissue\ %
ISSN 2150-8097. \\
\href{https://doi.org/\vldbdoi}{doi:\vldbdoi} \\
}\addtocounter{footnote}{-1}\endgroup

\ifdefempty{\vldbavailabilityurl}{}{
\vspace{.3cm}
\begingroup\small\noindent\raggedright\textbf{PVLDB Artifact Availability:}\\
The source code, data, and/or other artifacts have been made available at \url{\vldbavailabilityurl}.
\endgroup
}

\section{Introduction}
\label{sec:intro}
\begin{figure}[t!]
\centering
\begin{minipage}{\linewidth}
\centering
\includegraphics[width=\linewidth]{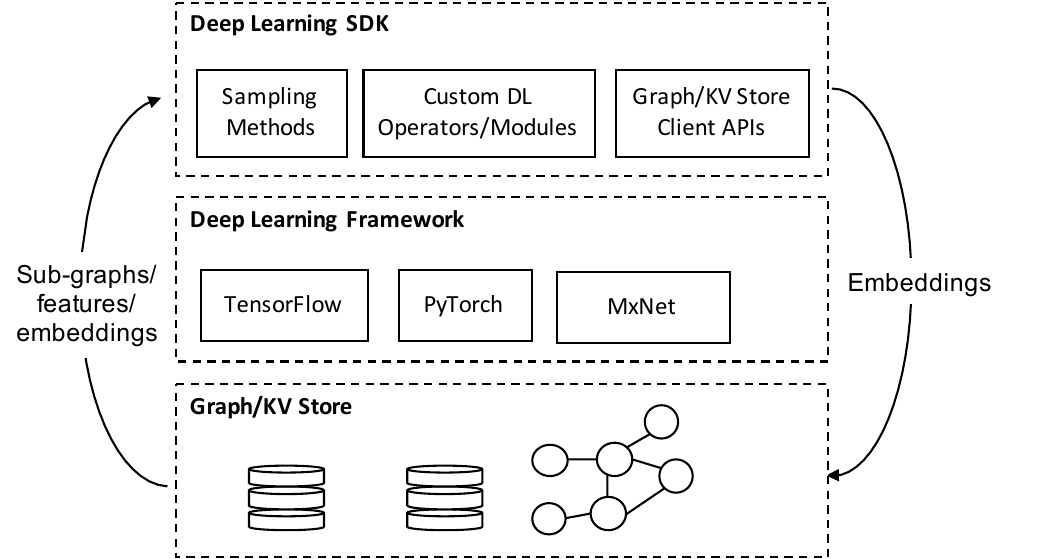}
\caption{A typical architecture of distributed frameworks.}
\label{fig:existing_solutions}
\end{minipage}
\begin{minipage}{0.8\linewidth}
\centering
\includegraphics[width=\linewidth]{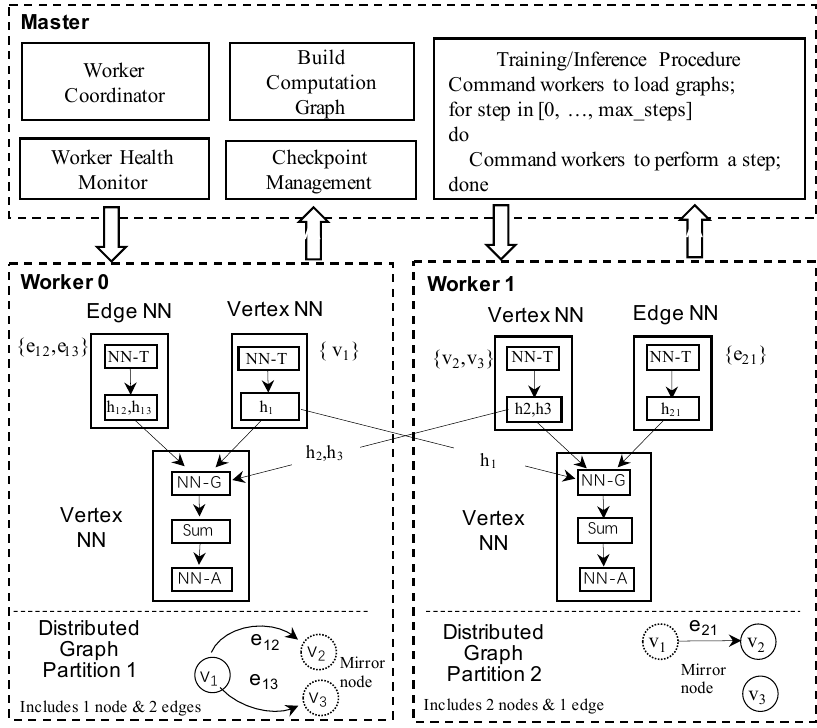}
\caption{The workflow of \projectname.}
\label{fig:graph_theta}
\end{minipage}
\vspace{-0.2cm}
\end{figure}

Graph neural networks (GNNs)~\cite{gori2005new, scarselli2008graph, scarselli2009the, ruiz2019gated, liu2019hyperboli, jia2020redundancy,zhang2021grain,cui2021metro} have been popularly used in analyzing graph data and achieved promising results in various applications, such as node and graph classification \cite{dai2016discriminative,yang2021heterogeneous,xu2018how,yu2022semisupervised}, link prediction \cite{lei2019gcn,zhang2018link}, program analysis~\cite{balog2017deepcoder, allamanis2018learning,liu2020unified}, recommendation~\cite{zhu2019aligraph,eksombatchai2018pixie,ying2018graph,yang2020session,li2021path} and quantum chemistry~\cite{gilmer2017neural}.
Currently, GNNs are implemented as proof-of-concept built upon deep learning (DL) frameworks like Tensorflow~\cite{abadi2016tensorflow}, PyTorch~\cite{paszke2019pytorch} and MXNet~\cite{chen2015mxnet}. 
They are either shared-memory~\cite{wang2019deep, eksombatchai2018pixie, ma2019neugraph, md2021distgnn, wan2022pipegcn, lin2020pagraph, mohoney2021marius, min2021large, yang2022gnnlab,liu2020g3} or use distributed computing~\cite{liao2018graph, zhu2019aligraph, fan2021graphscope, xu2021graphscope, zhang2020agl, jia2020improving, zheng2020distdgl, wang2021flexgraph, thorpe2021dorylus, gandhi2021p3, zheng2022distributed,zheng2022bytegnn}. The shared-memory frameworks are restricted to host memory size and difficult to handle big graphs. One approach to overcoming this constraint is to use external storage to store large graphs~\cite{mohoney2021marius}, and another is to use graph storage servers. Most of the distributed frameworks employ graph servers, which either
require every server to hold the entire graph or distribute the whole graph over a set of servers. In principle, the architecture of these distributed frameworks can be abstracted as Figure~\ref{fig:existing_solutions}, where classical graph computing and  MapReduce~\cite{dean2010mapreduce,zaharia2010spark} methods are usually used.
In addition, these distributed frameworks usually target only one type of training strategy.

For GNN training, global-batch~\cite{dai2016discriminative,kipf2016semi} and mini-batch~\cite{hamilton2017inductive} are two popular strategies.
Global-batch performs full graph convolutions across an entire graph by multiplying feature matrices with graph Laplacian. It requires calculations on the entire graph, regardless of graph density. In contrast, mini-batch conducts localized convolutions on a batch of subgraphs,
where a subgraph is constructed from a node with its neighbors. Typically, subgraph construction meets the challenge of size explosion, especially when the node degrees and the neighborhood exploration depth (neighborhood exploration depth is equal to the number of graph convolution layers) are very large~\cite{li2021training,chen2020simple,li2019deepgcns,xu2018representation,xu2021self}. 
This severely limits the scalability of mini-batch in processing dense graphs (dense graphs refer to graphs of high density) or sparse ones with highly skewed node degree distributions, and the situation further deteriorates very much for deep GNNs~\cite{li2021training,chen2020simple,li2019deepgcns,xu2018representation,xu2021self}.
Specifically, on one hand, not even mentioning dense graphs, high-degree nodes in a highly skewed graph can often cause a worker to become short of memory space to fully store a subgraph.
For example, in the Alipay dataset (see Section~\ref{sec:experiment_setups}), the degree of a node can reach hundreds of thousands.
On the other hand, a deep neighborhood exploration results in the explosion of a subgraph. 
In the Alipay dataset, a batch of subgraphs obtained from a two-hop neighborhood exploration with only 0.002\% nodes can generate a large size of 4.3\% nodes of the entire graph.
If the neighborhood exploration reaches 0.05\%, the batch of subgraphs reaches up to 34\% of the entire graph. In this way, one training step of this mini-batch will include 1/3 nodes in the forward and backward computation.
To alleviate the challenge of mini-batch, several neighbor sampling methods are proposed to reduce the size of neighbors~\cite{chen2018fastgcn,huang2018adaptive,ji2020accelerating,yoon2021performance}.
However, the sampling methods can bring unstable results during inference~\cite{eksombatchai2018pixie} and result in lower performance~\cite{jia2020improving,tripathy2020reducing, yoon2021performance}.

What is more important, in our industrial practice, we have been facing diverse demands from algorithm experts, including efficient processing of very large graphs, evaluation of different training strategies, support for deep GNNs, the effect of non-sampling, and even some combinations of them.
However,  no existing system is observed to be able to meet all of these requirements.
Overall, these aforementioned observations formulate the following key challenges we intend to address.
%
%
%

\textit{The first challenge is to support large skewed graphs with various sparsity.}
To address this challenge, we adopt a distributed graph training engine with customizable graph partitioning methods (refer to Sections~\ref{section:impl} and~\ref{result:partition}).
Along with this distributed engine, we implement a hybrid-parallel execution, in contrast with the conventional model-parallel and data-parallel training paradigms. This hybrid-parallel paradigm performs the forward and backward propagation computation of given graph data by a group of workers in a distributed fashion, thus enabling our system to scale up to big dense or sparse graphs.

\textit{The second challenge is to allow for conveniently exploring new training strategies in addition to the existing mini-batch and global-batch methods.}
For this challenge, existing tensor-based frameworks use sparse tensors\footnote{Sparse tensors are used to efficiently store and process tensors containing many zero values and typical formats include coordinate list (COO), compressed sparse row (CSR) and compressed sparse column(CSC).} to represent subgraphs and full graphs, which poses challenges to support the multiple training strategies described in this paper (especially global-batch) for the following reasons.
First, sparse tensors are nodes of the auto-differential computational graph and excessive sparse tensors will make the computational graph too large to be efficiently processed by the runtime of these frameworks, as described above.
Second, sparse tensors are with shared-memory and their sizes must be considerably constrained in local machines, thus infeasible for global-batch on large graphs.
To address global-batch, practitioners usually transform its computation to that of mini-batch.
Third, sparse tensors must carefully maintain the node/edge correspondence to the original graph (e.g. by some re-indexing techniques). 
In contrast,  we introduce a \textit{distributed subgraph abstraction} to transform these strategies into a unified computation (refer to Section~\ref{sec:sub_graph_training}).
The technical novelty of our distributed subgraph abstraction lies in that it unifies the processing of multiple training strategies and provides a consistent abstraction for the hybrid-parallel execution of distributed GNN training and inference. To the best of our knowledge, no existing system has ever made this.

\textit{The third challenge is to enable deep neighborhood exploration without neighbor sampling.}
To solve this challenge, with the help of our distributed subgraph abstraction built in the distributed graph training engine, neighborhood exploration only introduces a little extra storage overhead.
This overhead comes from the active set data structure that records the active status of nodes and edges, which can be proportional to the number of nodes or edges subject to implementations.
Therefore, sampling is no longer a necessity for the sake of reducing space complexity.
Furthermore, our distributed subgraph abstraction stores node/edge embedding (or activations) in place along with the graph topology, and thus naturally leads to distributed storage of embeddings (refer to Section~\ref{section:impl}).

To solve these challenges, we propose a new parallel and distributed GL system named \projectname, which is built from the ground up to support GNN training and inference and does not depend on existing DL frameworks like TensorFlow~\cite{abadi2016tensorflow} and PyTorch~\cite{paszke2019pytorch}.
The rationale behind such a design is explained as follows.
On one hand, graph operations can dominate the runtime~\cite{zheng2022bytegnn} and thus demand efficient graph processing infrastructures.
Built on top of our proprietary high-performance distributed graph processing system (also capable of serving existing DL frameworks), GraphTheta can keep seamless integration with our system and also makes its iterative development process independent of third-party DL frameworks.
On the other hand, the training procedure of DL can be modeled as an auto-differentiable computational graph of data and operator (or neural network function) nodes.
For each training step, we can build the computational graph by means of treating vertices/edges as data nodes and then building data-operator or operator-operator linkages following the GNN and the input graph topology.
In this case, large graphs will make the computational graph inefficiently processed by the runtime of existing frameworks.
Subject to specific implementations, the size of the computation graph can have a higher order of magnitude than the input graph.
Figure~\ref{fig:graph_theta} illustrates the workflow of our system.
There are two roles in our system, i.e. master and worker. The master process coordinates the execution of all workers, monitors their health, manages checkpoints, and directs the learning procedure including training and inference.
The workers wait and receive commands from the master, and then execute them.
The communication between the master and workers is done using remote procedure calls.
\begin{algorithm}[t!]
\scriptsize
\caption{MPGNN under different training strategies.}
\label{alg:cbgnn}
\begin{algorithmic}[1]
\STATE Partition graph $\mathcal{G}$ into $P$ non-intersect subgraphs, denoted as $\mathcal{G}=\bigcup\{\mathcal{G}_{1},\mathcal{G}_{2},...,\mathcal{G}_{P}\}$. 
\STATE $\mathcal{G}_{p} \leftarrow \mathcal{G}_{p}\cup {N}_K(\mathcal{G}_{p}) \;\; \forall p \in [1,P]$
\FOR{each $r \in [1,St]$}
\STATE{Pick $\gamma$ subgraphs randomly, $\mathcal{B}_r \leftarrow \bigcup_{\gamma}\{\mathcal{G}_{i}\}$}
\FOR{each $k \in [1,K]$}
\STATE $\bm{n}_i^k \leftarrow \textit{Proj}_k(\bm{h}_i^{k-1}|\bm{W}_k)\;\; \forall v_i \in \mathcal{B}_r$
\STATE $\bm{m}_{j\rightarrow i}^k  \leftarrow \textit{Prop}_k\Big(\bm{n}_{j}^{k},\bm{e}_{i,j},\bm{n}_{i}^{k}\Big|\bm{\theta}_{k}\big)\;\;\forall e_{i,j} \in \mathcal{B}_r$
\STATE $\bm{h}_i^{k} \leftarrow \textit{Agg}_k\Big(\bm{h}_{i}^{k-1},\big\{\bm{m}_{j\rightarrow i}^k\big\}_{j \in N_O(i)} \big| \bm{\mu}_{k} \Big)\;\; \forall v_i \in \mathcal{B}_r$
\ENDFOR
\STATE $\hat{\bm{y}}_i \leftarrow \textit{Dec}(\bm{h}_{i}^K\big|\bm{\omega}) \; \forall v_i \in \mathcal{B}_r$
\STATE ${L}_r \leftarrow \sum_{v_i \in \mathcal{B}_r}l(\bm{y}_i,\hat{\bm{y}}_i)$
\STATE Update parameters by gradients of ${L}_r$
\ENDFOR
\end{algorithmic}
\end{algorithm}

We have conducted extensive experiments on various networks to evaluate the performance of our system. 
Our experimental results revealed that \projectname can learn GNNs as well as existing frameworks in terms of model accuracy.
Furthermore, \projectname scales well to 1,024 CPU workers in our production Kubernetes~\cite{kubernetes2022} CPU cluster and can train an in-house developed GNN on the Alipay dataset of 1.4 billion nodes and 4.1 billion attributed edges using three training strategies.
To our knowledge, this is the largest GNN learning task that uses edge attributes of a billion-scale industrial network in the literature\footnote{The network in \cite{ying2018graph} has 3 billion nodes and 18 billion edges with no edge attributes.}. 
Furthermore, \projectname outperforms DistDGL both in terms of training speed (up to $2.02\times$ faster on the same machines) and scalability, and can run $2.61\sim 30.56\times$ faster than GraphLearn (the open-source version of AliGraph~\cite{zhu2019aligraph}).
Our technical contributions are summarized as follows. 
\begin{itemize} [leftmargin=*]
\item We present a distributed GL system implemented based on a vertex-centric graph processing programming model. This system proposes a new distributed graph training engine developed with gradient propagation support. This engine implements a new hybrid-parallel execution and supports GNN training and inference via a unified implementation. Compared to existing data-parallel executions, the new engine can scale up to big dense/sparse graphs and enable deep neighborhood exploration.
\item \projectname introduces a new GL abstraction NN-TGAR~(\underline{N}eural \underline{N}etwork \underline{T}ransform-\underline{G}ather-\underline{A}pply-\underline{R}educe), which enables user-friendly programming (supporting distributed training) and bridges the gap between graph processing paradigms and deep learning computational patterns.
This abstraction centers around the semantics of nodes and edges, instead of tensors as in conventional DL frameworks, and is used to compose DNNs by orchestrating operations on nodes and edges.
\item To alleviate the redundant calculation among batches, \projectname supports a new type of training strategy, i.e. cluster-batch~\cite{ClusterGCN}, which performs graph convolution on a cluster of nodes and can be taken as a generalization of mini-batch or global-batch. In particular, for each training strategy, our system scales well to 1,024 workers on the billion-scale industrial Alipay dataset in a cluster of CPU virtual machines of small memory each.
\end{itemize}

\section{Preliminary}
\label{GNN}

\subsection{Notations}
A graph can be defined as $\mathcal{G = \{V, E\}}$,
where $\mathcal{V}$ and $\mathcal{E}$ are nodes and edges respectively.
For simplicity, we define $N=|\mathcal{V}|$ and $M=|\mathcal{E}|$.
Each node \mbox{$v_i \in \mathcal{V}$} is associated with a feature vector $\bm{h}_i^0$, 
and each edge $e(i,j) \in \mathcal{E}$ has a feature vector  $\bm{e}_{i,j}$, a weight value $a_{i,j}$.
To represent the structure of $\mathcal{G}$, we use an adjacency matrix \mbox{$\bm{A} \in \mathbb{R}^{N\times N}$}, where \mbox{$A(i,j) = a_{i,j}$} if there exists an edge between nodes $i$ and $j$, otherwise $0$.
\subsection{Graph Neural Networks}
\label{sec:gnns}
Existing graph learning tasks are often comprised of \texttt{encoders} and \texttt{decoders} as stated in~\cite{hamilton2017representation}.
Encoders map high-dimensional graph information into low-dimensional embeddings.
Decoders are application-driven. The essential difference between GNNs relies on the encoders.
Encoder computation can be performed by two established methods~\cite{zhang2020deep}: spectral and propagation.
The spectral method generalizes convolution operations on two-dimensional grid data to graph-structured data as defined in~\cite{kipf2016semi,NIPS2016_6081}, which uses sparse matrix multiplications to perform graph convolutions.
The propagation method describes graph convolutions as a message propagation operation, which is equivalent to the spectral method (refer to Section~\ref{equivalence_relation} in the supplementary material\footnote{https://github.com/yongchao-liu/graphtheta} for the proof). 

In this paper, we present an algorithmic framework of Message Propagation based Graph Neural Network (MPGNN) as a typical use case to elaborate the compute pattern of NN-TGAR and our new system \projectname.
As shown in Algorithm~\ref{alg:cbgnn},  MPGNN can unify existing GNN algorithms under different training strategies.
Recent work~\cite{2017GAT, hamilton2017inductive, MPNN} focuses on propagating and aggregating messages and aims to deliver a  general framework for different message propagation methods.
The core difference between existing propagation methods lies in the projection (Line~6), the message propagation (Line~7), and the aggregation (Line~8) functions.

\subsection{Cluster-batched Training} 
\label{community_batch}
We study a new training strategy, namely cluster-batched gradient descent, to address the challenge of redundant neighbor embedding computation under the mini-batch strategy. 
This training strategy was first used by the Cluster-GCN~\cite{ClusterGCN} algorithm and shows superior performance to mini-batch in some applications.
This method can maximize per-batch neighborhood sharing by taking advantage of community detection (graph clustering) algorithms.

Cluster-batch first partitions a big graph into a set of smaller clusters and then generates a batch of data either based on one cluster or a combination of multiple clusters. 
Similar to mini-batch, cluster-batch also performs localized graph convolutions. However, cluster-batch restricts the neighbors of a target node into only one cluster, which is equivalent to conducting a full graph convolution on a cluster of nodes.
Typically, cluster-batch generates clusters by using a community detection algorithm based on maximizing intra-community edges and minimizing inter-community connections~\cite{louvain}.
Note that community detection can run either beforehand or at runtime, based on the requirements. Moreover, cluster sizes are often irregular, resulting in varied batch sizes.
It needs to be noted that differing from the cluster-batch in Cluster-GCN, our cluster-batch optionally allows for target nodes to access boundary neighbors outside the clusters.
In the supplementary material, Figure~\ref{fig:batchexample} illustrates an example for cluster-batched computation and Table~\ref{tbl:comparison} compares the pros and cons of the three training strategies, implying the necessity to design a new GNN learning system that enables the exploration of different training strategies and a solution to address the limitations of existing architectures.

\section{Compute Pattern}

\subsection{NN-TGAR}
To solve the GNN learning in a hybrid-parallel fashion, we present a general computation pattern abstraction, namely \texttt{NN-TGAR},
which can perform the forward and backward computation of GNNs on a big (sub-)graph distributively.
This abstraction decomposes an encoding layer (explained in the context of encoders and decoders as described in Section~\ref{sec:gnns}) into a sequence of independent stages, i.e., \texttt{NN-Transform}~(\texttt{NN-T}), \texttt{NN-Gather}~(\texttt{NN-G}), \texttt{Sum}, \texttt{NN-Apply}~(\texttt{NN-A}), and \texttt{Reduce}. Our method allows distributing the computation over a cluster of machines.
\texttt{NN-T} is executed on each node to transform the results and generate messages.
\texttt{NN-G} is applied to each edge, where the inputs are edge values, source node, destination node, and the messages generated in the previous stage.
After each iteration, this stage updates edge values and sends messages to the destination node (maybe on other workers).
In \texttt{Sum}, each node accumulates the received messages by means of a non-parameterized method like averaging, concatenation or a parameterized one such as \texttt{LSTM}~\cite{liu2019geniepath}.
The resulting summation is updated to the node by \texttt{NN-A}.

Different from the vertex-centric programming abstraction Gather-Sum-Apply-Scatter~(\texttt{GAS}) proposed by PowerGraph~\cite{gonzalez2012powergraph} for conventional graph processing applications,
\texttt{NN-T}, \texttt{NN-G} and \texttt{NN-A} are implemented as neural network functions.
In forwards, the trainable parameters also join in the calculation of the three stages, but kept unchanged.
In backwards, the gradients of these parameters are generated in \texttt{NN-T}, \texttt{NN-G} and \texttt{NN-A} stages,
which is used in the final stage \texttt{NN-Reduce} for parameter updating.
\texttt{NN-TGAR} can be executed either on the entire graph or subgraphs, subject to the training strategy used.

Intuitively, we can understand \texttt{NN-TGAR} as follows. If having only operations on individual nodes (e.g. imagining a node is an image), an NN can be realized only by \texttt{NN-T}. The forward and backward computation logics are almost identical to the NNs in the tensor-based DL frameworks. If needing message passing along with edges, \texttt{NN-G}, \texttt{Sum} and \texttt{NN-A} will be used.
If a node aggregates information from its neighbor along every out-edge in the forward computation, this node will aggregates gradient from its neighbor along every in-edge in the backward, and vice versa.
\begin{figure*}[t]
\setlength{\abovecaptionskip}{0.1cm}
	\centering
	\subfigure[Forward from $h_i^{k-1}$ to $h_i^{k}$: \texttt{NN-T} uses the $k$th projection function, \texttt{NN-G} uses the $k$th propagation function and \texttt{NN-A} uses the $k$th aggregation function]{
		\begin{minipage}[b]{0.36\linewidth}
			\centering
			\includegraphics[width=\linewidth]{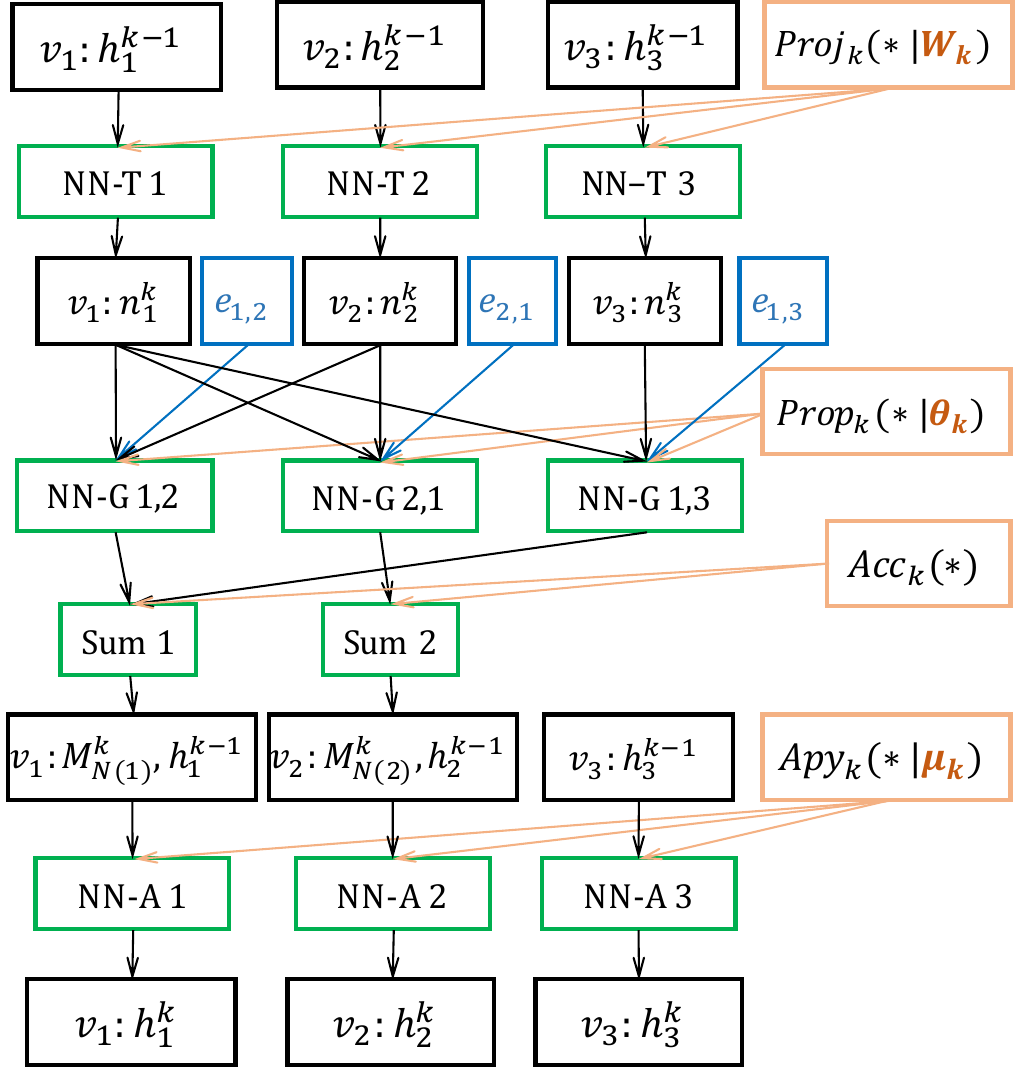}
			\label{fig:forward}
		\end{minipage}%
	}%
	\quad
	\subfigure[Backward from $\nabla h_i^{k}$ to $\nabla h_i^{k-1}$: \texttt{NN-T} uses the derivation of $Apy_k$, \texttt{NN-G} uses the deriv. of $Acc_k\&Prop_k$, \texttt{NN-A} uses the deriv. of $Proj_k$, and \texttt{NN-R} processes gradients of parameters.]{
		\begin{minipage}[b]{0.402\linewidth}
			\centering
			\includegraphics[width=\linewidth]{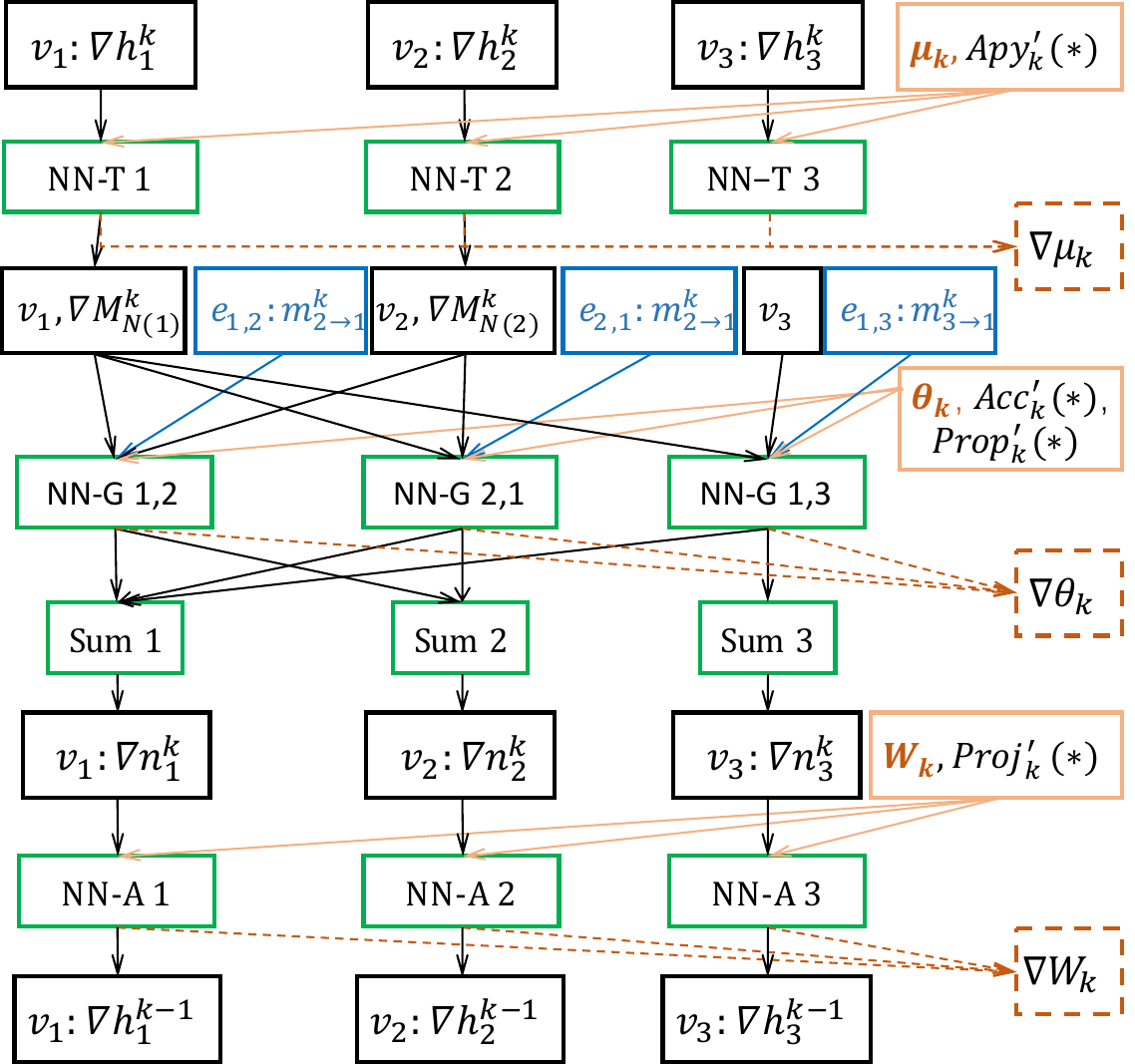}
			\label{fig:backward}
		\end{minipage}%
	}%
	\quad 
	\subfigure[The example graph used in (a$\backslash$b), nodes may on different workers.]{
		\begin{minipage}[b]{0.148\linewidth}
		\centering
		\includegraphics[width=.75\linewidth]{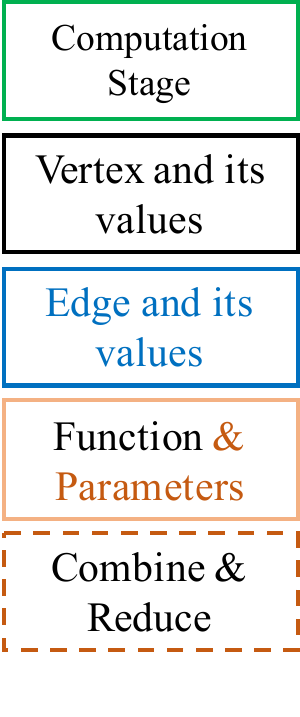}
		\centering
		\includegraphics[width=\linewidth]{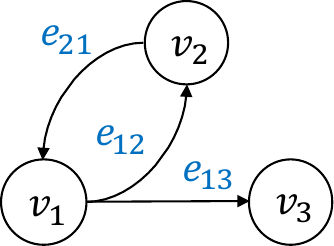}
		\label{fig:demo_graph}
		\end{minipage}%
	}%
\vspace{-0.3cm}
\caption{The computation pattern of NN-TGAR.}
\label{compute_process}
\vspace{-0.4cm}
\end{figure*}
\subsection{Forward}
The forward of a GNN model can be described as $K+2$ passes of \texttt{NN-TGA},
as each encoding layer can be described as one pass of \texttt{NN-TGA}. 
The decoder and loss functions can be separately described as a single \texttt{NN-T} operation.
Recall that the decoder functions are application-driven. A decoder function can be described by a single \texttt{NN-T} operation in node classification, and a combination of \texttt{NN-T} and \texttt{NN-G} in link prediction.
Without loss of generality, we use node classification as the default task in this paper.
The forward is comprised of $K$ passes of \texttt{NN-TGA} from the first to \textit{K}-th encoding layer, one \texttt{NN-T} operation for the decoder function, and one for the loss calculation.
In the forward of each encoding layer, the projection function is executed on \texttt{NN-T} and the propagation function on \texttt{NN-G}.
However, the aggregation function is implemented by a combination of \texttt{Sum} and \texttt{NN-A}, which corresponds to the accumulate part and the apply part defined as follows: 
\begin{subequations}
\label{eq:aggregation}
\begin{equation}
\label{eq:acc}
\bm{M}_{i}^k=\textit{Acc}_k\big(\big\{\bm{m}_{j\rightarrow i}^k\big\}_{j \in N(v_i)} \big| \bm{\mu}_k^{(1)}\big),
\end{equation}
\begin{equation}
\label{eq:apy}
\bm{h}_i^k=\textit{Apy}_k\big(\bm{h}_i^k,\bm{M}_{N_O(i)}^k \big| \bm{\mu}_k^{(2)}\big),
\end{equation}
\end{subequations}
where $\bm{\mu}_k=\big[\bm{\mu}_k^{(1)},\bm{\mu}_k^{(2)}\big]$.
If the accumulate part is not parameterized by mean-pooling, 
$\bm{\mu}_k^{(1)} = \bm{0}$ and  $\bm{\mu}_k = \bm{\mu}_k^{(2)}$.

As shown in Figure~\ref{fig:forward},
the embedding of the ($k-1$)-th layer $\bm{h}_i^{k-1}$ is transformed to $\bm{n}_i^{k}$ using a neural network $\textit{Proj}_k(*|\bm{W}_k)$ for all the three nodes in \texttt{NN-T}.
The \texttt{NN-G} stage collects messages both from the neighboring nodes $\{v_2, v_3\}$ (and $v_1$) 
and from adjacent edges $\{e_{1,2}, e_{1,3}\}$ (and $e_{2,1}$) to the centric node $v_1$ (and $v_2$) through the propagation function $\textit{Prop}_k(*|\theta_k)$.
The output of this stage is automatically accumulated by $\textit{Acc}_k(*)$, resulting in the summed message $\bm{M}_{N(1)}^k$ and $\bm{M}_{N(2)}^k$.
The \texttt{NN-A} stage computes the new embeddings of nodes $\{v_1, v_2, v_3\}$, i.e. $\bm{h}_1^{k}$, $\bm{h}_2^{k}$ and $\bm{h}_3^{k}$, as the output of this encoding layer.

\subsection{Auto-Differentiation and Backward}
Auto-differentiation is a prominent feature in deep learning and \projectname also implements auto-differentiation to simplify GNN programming by automating the backward gradient propagation during training.
Like existing deep learning training systems, a primitive operation has two implementations: a forward version and a backward one.
For instance, a data transformation function can implement its forward computation by a sequence of primitive operations.
In this case, its backward computation can be automatically interpreted as a reverse sequence of the backward versions of these primitive operations.
Assuming that $\bm{y}=f(\bm{x},\bm{W})$ is a user-defined function composed of a sequence of built-in operations,
\projectname can generate the two derivative functions automatically: $\partial \bm{y}/\partial \bm{x}=f^{\prime}_{x}(\bm{W})$ and $\partial \bm{y}/\partial \bm{W}=f^{\prime}_{W}(\bm{x})$.
\texttt{NN-TGAR} will organize all these derivative functions to implement the backward progress of the whole GNN model.
Refer to Section~\ref{backwards_gnn} in supplementary material for a general proof of backward computation with message propagation as well as Section~\ref{backwards_tgar}
for the derivatives of MPGNN.

In the task of node classification, the backward of a GNN model can be described as $K+2$ passes of  \texttt{NN-TGAR}, but  in a reverse order.
First, the differential of a loss function $\partial{L}/\partial\hat{\bm{y}}_i=l^{\prime}(\bm{y}_i)$ is executed on each labeled node by a single \texttt{NN-T} stage.
Then, the two stages \texttt{NN-T} and \texttt{Reduce} are used to the calculate the differential of the decoder function.
In this phase, the gradients of the final embedding for each node are calculated as 
$\partial L/ \partial \bm{h}_{i}^K=\partial L/\partial \hat{\bm{y}}_i\cdot\textit{Dec}^{\prime}(\bm{\omega})$
and updated to the corresponding node.
Meanwhile, the gradients of the decoder parameters are calculated as
$\partial L/ \partial \bm{\omega}=\partial L/\partial \hat{\bm{y}}_i\cdot\textit{Dec}^{\prime}(\bm{h}_{i}^K)$
and sent to the optimizer.
The $K$ passes of \texttt{NN-TGAR} are executed backwards from the \textit{K}-th to the first encoding layer.

Differing from forward that runs the apply part at the last step,  backward executes the differential of the apply part on nodes $\{v_1, v_2, v_3\}$ in the \texttt{NN-T} stage, as shown in Figure~\ref{fig:backward}.
This stage calculates $\partial L/ \partial \bm{\mu}^k$ that will be sent to the optimizer, $\{\partial L/ \partial \bm{h}_{i}^{k-1}\cdot\partial\textit{Apy}_{k}/\partial \bm{n}_i^k | i=1, 2, 3\}$ that will be updated to the nodes as values, and \mbox{$\{\partial L/ \partial \bm{M}_{i}^k | i = 1, 2\}$} that will be consumed by the next stage \texttt{NN-G}.
Besides receiving messages from the previous stage, stage \texttt{NN-G} takes as input the values of the corresponding adjacent edges and centric nodes, and sends the result to neighbors and centric nodes, as well as the optimizer.

In \texttt{NN-G}, taking $\texttt{Gather}_{1,3}$ as an example, the differential of the accumulate function calculates $\partial L/ \partial \bm{m}_{3\rightarrow1}^k$. Subsequently, the differential of the propagation function computes $\partial \textit{Prop}_k / \partial \bm{n}_3^k$, $\partial \textit{Prop}_k / \partial \bm{n}_1^k$, and $\partial \textit{Prop}_k / \partial \bm{\theta}_k$, all of which are multiplied by $\partial L/ \partial \bm{m}_{3\rightarrow1}^k$ and then sent to the source node $v_3$, destination node $v_1$, and the optimizer.
The \texttt{Sum} stage receives gradient vectors computed in \texttt{NN-G}, as well as new node values computed in \texttt{NN-T}, and then element-wisely adds the gradients by node values for each of the three nodes, resulting in \mbox{$\{\partial L/ \partial \bm{n}_i^k | i = 1, 2, 3\}$}.
These three results will be passed to the next stage.
The \texttt{NN-A} computes the gradients of the ($k-1$)-th layer embeddings $\partial L/ \partial \bm{h}_i^{k-1}$ as $\partial L/ \partial \bm{n}_i^k \cdot \textit{Proj}_k^{\prime}(\bm{W}_k)$, where the gradients of $\bm{W}_k$ are calculated similarly.
The optimizer invokes \texttt{Reduce} to aggregate all the gradients of parameters (i.e., $\bm{\mu}_k$, $\bm{\theta}_k$, and $\bm{W}_k$), which are generated in stages \texttt{NN-T}, \texttt{NN-G} and \texttt{NN-A} and distributed over nodes/edges, and updates the parameters with this gradient estimation.

\section{Implementation}
\label{section:impl}
\begin{figure}[t!]
\centering
\begin{minipage}{\linewidth}
\centering
\includegraphics[width=0.8\linewidth]{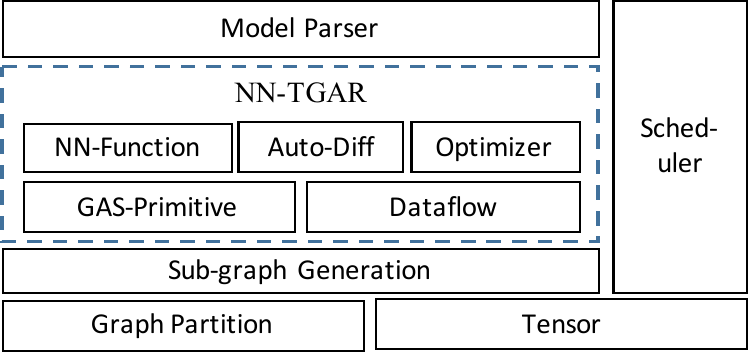}
\caption{The system architecture of \projectname.}
\label{arch_fig}
\end{minipage}
\end{figure}

Inspired by distributed graph processing systems, we present a new GNN training system which can simultaneously support all of the three training strategies. 
Our new system enables deep GNN exploration without pruning graphs. 
Moreover, our system can balance memory footprint, time cost per epoch, and convergence speed.
Figure~\ref{arch_fig} shows the architecture of our system. 
It consists of five components: 
($i$) a graph storage component with distributed partitioning and heterogeneous features and attributes management, 
($ii$) a subgraph generation component with sampling methods, 
($iii$) graph operators which manipulate nodes and edges, 
and ($iv$) learning core operations including neural network operators (including fully-connected layer, attention layer, batch normalization, concat, mean/attention pooling layers and etc.), typical loss functions (including softmax cross-entropy loss and binary cross-entropy loss), and optimizers (including SGD, Adam~\cite{kingma2014adam} and AdamW~\cite{loshchilov2018decoupled}).


\subsection{Distributed Graph Representation}
Our graph programming abstraction follows the vertex-program paradigm and fits the computational pattern of GNNs which centers around nodes. In our system, the underlying graphs are stored distributively which require fast graph partitioning algorithms for efficient processing. A number of graph partitioning approaches have been proposed, such as vertex-cut~\cite{ccatalyurek1996decomposing, gonzalez2012powergraph, jain2013graphbuilder}, edge-cut~\cite{karypis1999parallel, stanton2012streaming, tsourakakis2014fennel}, and hybrid-cut~\cite{chen2019powerlyra} solutions. Typical graph partitioning algorithms are included in our system to support popular graph processing methods (The effect of different partitioning methods on performance will be elaborated in Section~\ref{result:partition}).

\begin{figure*}[t!]
\subfigure[]{
\begin{minipage}[b]{0.35\linewidth}
\includegraphics[width=\linewidth]{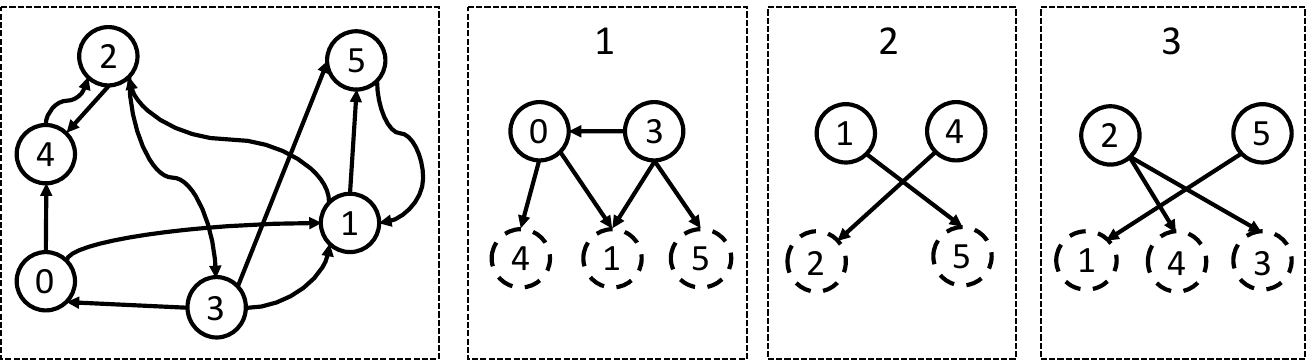}
\vspace{-0.6cm}
\label{partition_fig}
\end{minipage}
}
\subfigure[]{
\begin{minipage}[b]{0.4\linewidth}
\includegraphics[width=\linewidth]{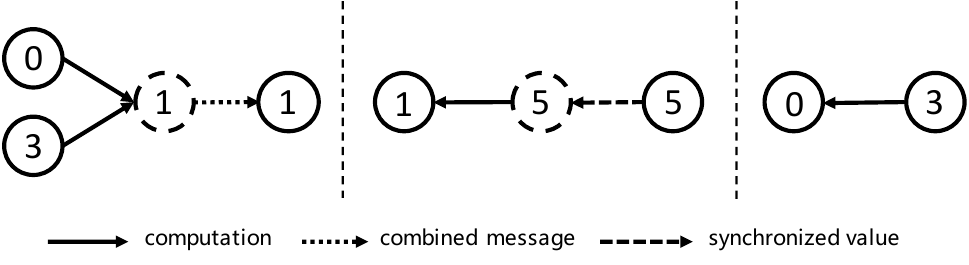}
\vspace{-0.6cm}
\label{message_fig}
\end{minipage}
}
\subfigure[]{
\begin{minipage}[b]{0.2\linewidth}
\includegraphics[width=\linewidth]{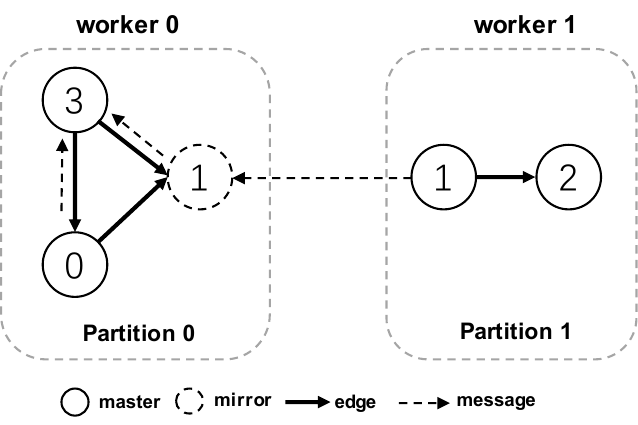}
\vspace{-0.6cm}
\label{fig:traversal}
\end{minipage}
}
\vspace{-0.3cm}
\caption{Three examples: (a) partitioning nodes evenly with mirrors denoted by dotted lines; (b) computation and communication in the \texttt{Gather} primitives: compute, combine and synchronized; and (c) distributed graph traversal.}
\vspace{-0.6cm}
\end{figure*}

To efficiently run GNN-oriented vertex programs, we propose a new distributed graph representation format which evenly distributes nodes to partitions and cuts off cross-partition edges. 
Similar to PowerGraph, we use master and mirror nodes, where a master node is assigned to one partition and its mirrors are created in other partitions. 
For each edge, our method assigns it to the partition in which its source node is a master (target nodes also can be used as the indicator).
This way, any edge contains at least one master node.

The vertex-cut approach used by PowerGraph has the disadvantage of duplicating mirror nodes, resulting in memory overhead for multi-layer GNN learning.
To address this problem, our method allows mirror nodes to act as placeholders and only hold node states instead of the actual values.
With this strategy, our method not only reduces memory overhead but also lowers communication overhead for two reasons. On one hand, by allowing mirror nodes not to hold actual values, our method can significantly reduce memory overhead. On the other hand, at the end of each superstep, PowerGraph will synchronize all master nodes to their mirrors. Instead, our method removes this global synchronization,  and only synchronizes the masters used. This means that the communication overhead of PowerGraph is an upper bound of ours.
Moreover, our method can reduce the replica factor to 1 from $(N_{master} + N_{mirror})/N_{master}$, where $N_{master}$ and $N_{mirror}$ are the number of master and mirror nodes. 
With the partitioning method, the implementation of each primitive in \texttt{GAS} abstraction is composed of several phases.
Figure~\ref{message_fig} illustrates the computation of the \texttt{Gather} primitive. 

To traverse the graph efficiently, \projectname organizes outgoing edges in CSR and incoming edges in CSC, and stores node and edge values separately.
Our distributed graph traversal is completed in two concurrent operations: one traverses nodes with CSR and the other with CSC.
For the operation with CSR, each master node sends its related values to all the mirrors and then gathers its outgoing edges with master neighbors, where the edges with mirror neighbors are directly skipped.
For the operation with CSC, each mirror node gathers incoming edges with master neighbors.
Mirror nodes receiving values from their corresponding masters will be passively gathered by their neighbors.

Figure~\ref{fig:traversal} illustrates an example of traversing all outgoing edges of the master node 3 in partition 0 held by worker 0.
For the operation with CSR, worker 0 processes the outgoing edge of node 3 to the master node 0.
As node 1 in worker 0 is a mirror, the edge from node 3 to node 1 is skipped.
Meanwhile, worker 1 sends the value of its master node 1 to the mirror in worker 0.
Concurrently, for the operation with CSC, worker 0 processes the incoming edge of mirror node 1 from node 3.
Herein, the mirror node 1 receives its value sent from worker 1 and will be passively gathered by node 3 in worker 0.
Overall, we can see that communication only occurs between master and mirror nodes.

The abstraction can address the local message bombing problem. 
On one hand, for a master-mirror pair, we only need one time of message propagation of node values and the results, which can reduce the traffic load from $O(M)$ to $O(N)$. 
This is because in a partition, a master/mirror node is shared by all of its edges located in the partition, and our method that merely synchronizes master nodes with mirrors makes the traffic load proportional to $N$ rather than $M$.
On the other hand, we only synchronize node values involved in the computation per neural network layer, during the dataflow execution of a GNN model. 
Moreover, we remove the implicit synchronization phase after the original \texttt{Apply} primitive, instead of introducing implicit master-mirror synchronization to overlap computation and communication. 
Additionally, heuristic graph partitioning algorithms, such as~METIS~\cite{metis} and Louvain~\cite{louvain}, are also supported to adapt cluster-batched training.

\subsection{Subgraph Training}
\label{sec:sub_graph_training}
To unify the processing of all three types of training strategies, our new system uses subgraphs as the abstraction of graph structures and the GNN operations are applied.
Both mini-batch and cluster-batch train a model on the subgraphs generated from the initial batches of target nodes, whereas global-batch does on the entire graph.
The size of a subgraph can vary from one node to the entire graph based on three factors, i.e., the number of GNN layers, graph topology, and community detection algorithms.
The number of GNN layers determines the neighbor exploration depth and has an exponential growth of subgraph sizes.
For graph topology, node degree determines the exponential factor of subgraph growth.

\begin{figure}[!t]
\centering
\begin{minipage}{\linewidth}
\centering
\includegraphics[width=0.95\linewidth]{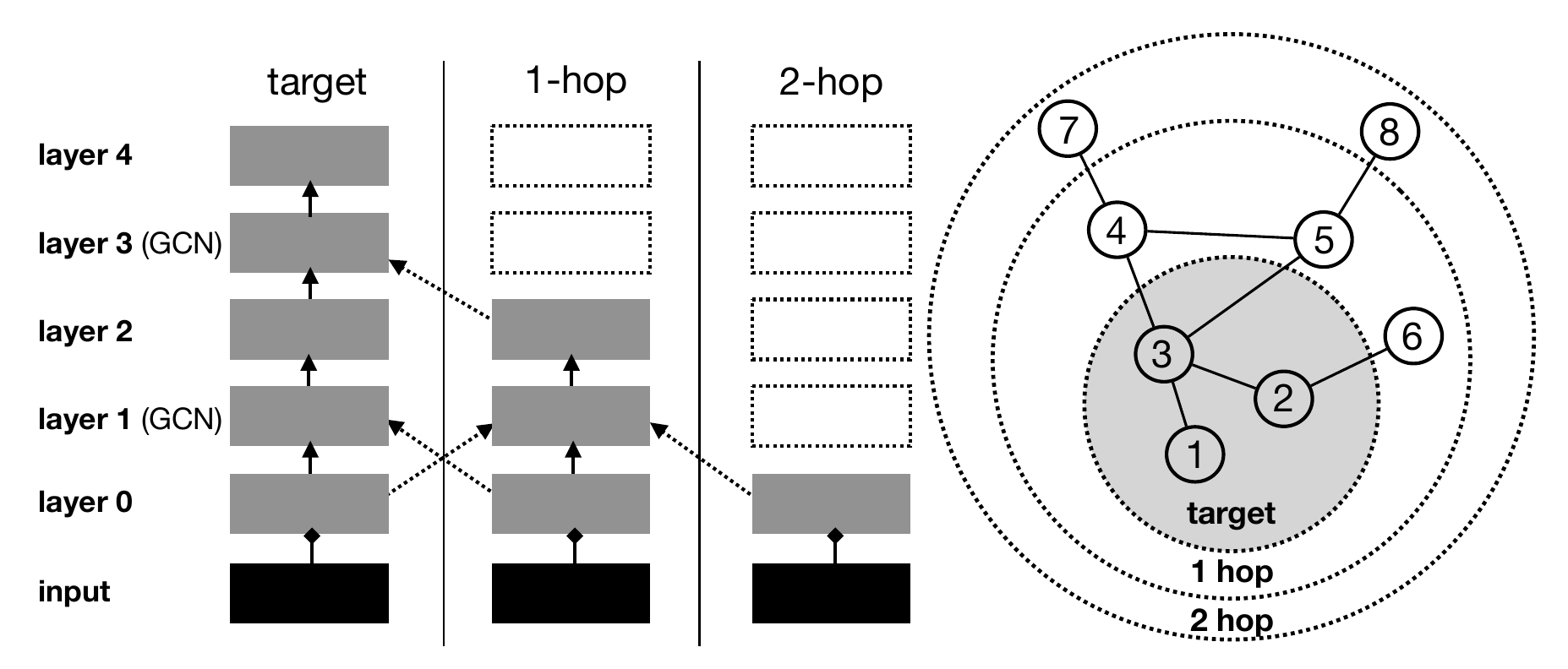}
\caption{The tensor stacks of a target node and its 2-hop neighbors in the training process with a 2-hop GNN model.}
\label{trainingtarget_fig}
\end{minipage}
\begin{minipage}{\linewidth}
\centering
\includegraphics[width=0.95\linewidth]{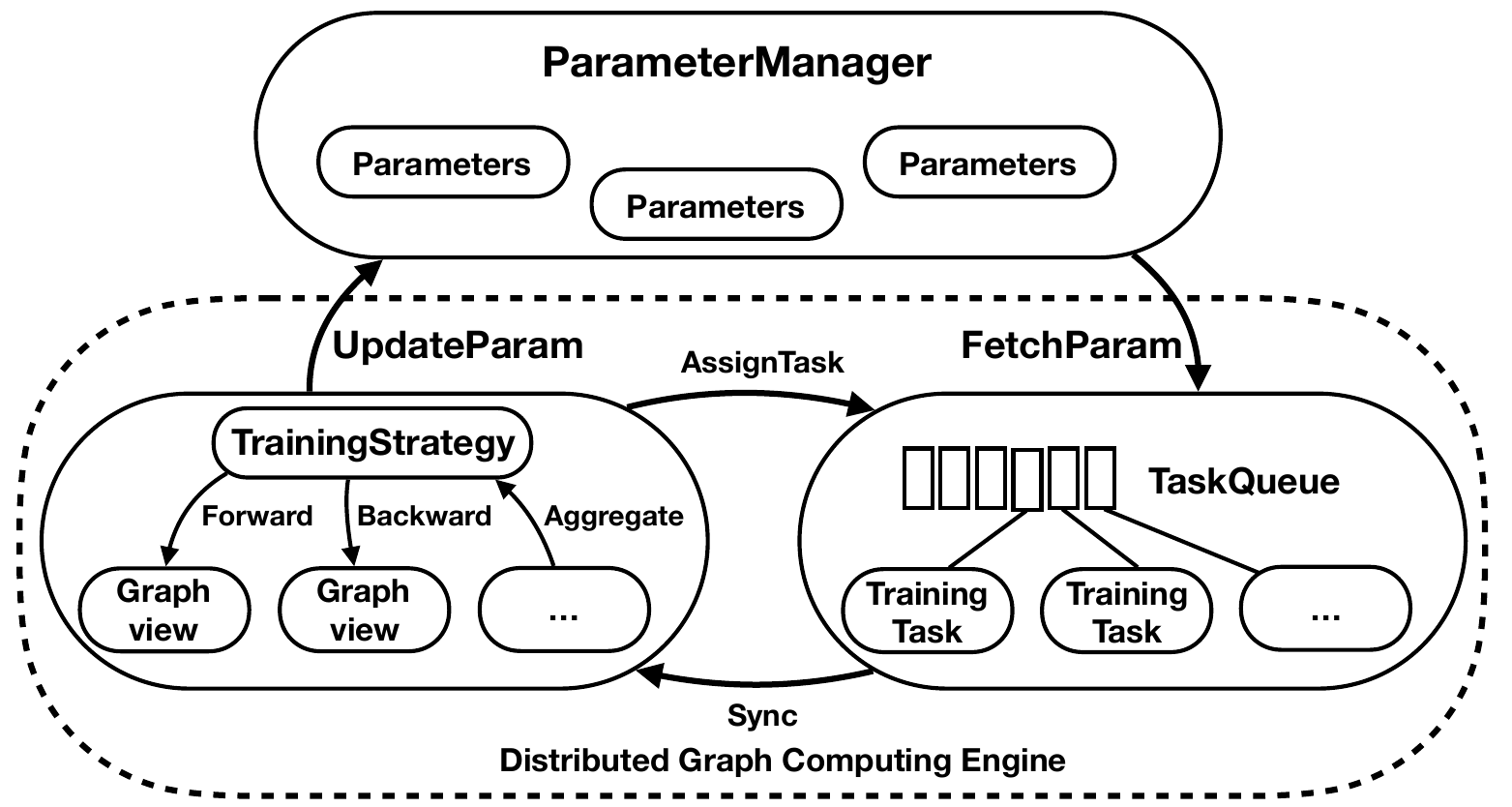}
\caption{The parallel batched training paradigm.}
\label{paralleltraining_fig}
\end{minipage}
\vspace{-0.15cm}
\end{figure}

Figure~\ref{trainingtarget_fig} shows a mini-batch training example, where a GNN model containing two graph convolution layers is interleaved by two fully-connected layers.
The right part shows a subgraph constructed from a batch of initial target nodes \mbox{$\{1, 2, 3\}$}, which has one-hop neighbors \mbox{$\{4, 5, 6\}$} and two-hop neighbors \mbox{$\{7, 8\}$}.
The left part shows the forward and backward computation of this subgraph, with arrows indicating the propagation direction.
For subgraph computation, a straightforward method is to load the subgraph structure and the related data into memory and perform matrix/tensor operations on the subgraph located on the same machine.
As the memory overhead of a subgraph may exceed the memory limit, this method has an inherent limitation in generalization.
Instead, \projectname completes the training procedure with distributed computing by only spanning the structure of the distributed subgraph.
More specifically, our system distributively stores the input and intermediate data of nodes/edges following the distributed subgraph, instead of transporting and placing all of them into a shared memory space.

To construct a subgraph, our system introduces a breadth-first-search traversal operation.
For each target node, this operation initializes a minimal number of layers per node, which are involved in the computation, to reduce unnecessary propagation of graph computing.
Furthermore, to avoid the cost of subgraph structure construction and preserve graph access efficiency, we build a vertex-ID mapping between the subgraph and the local graph within each process/worker to reuse CSR/CSC indexing.
In addition, our system has implemented a few sampling methods, including random neighbor sampling~\cite{hamilton2017inductive}, which can be applied to subgraph construction.

\subsection{Parallel Execution Model}
\label{sec:execution_model}

Training subgraphs sequentially cannot fully unleash the power of a distributed system.
\projectname is designed to concurrently train multiple subgraphs with multi-versioned parameters in a distributed environment, 
and support concurrent lookups and updates, which lead to two distinguished features from the existing GNN training systems: ($i$) parallel subgraph tensor storage built upon distributed graphs, and ($ii$) \texttt{GraphView} abstraction and multi-versioned parameter management to enable parallelized batched training.

\textbf{Parallel tensors storage.} For subgraph training, we investigate two key techniques to enable low-latency access to distributed subgraphs and lower total memory overhead.
The first technique is reusing the CSR/CSC indexing.
It is inefficient in a high-concurrency environment to construct and release the indexing data for each subgraph on the fly.
Instead, the global indexing of the whole graph is reused, 
and a private cache-friendly vertex-ID mapping is adopted to efficiently access a graph topology, as described in section~\ref{sec:sub_graph_training}.

The second technique is task-oriented tensor storage.
In GNNs, the same node can be incorporated in different subgraphs constructed from different batches of target nodes, especially for the nodes of high degrees.
To make tasks context-independent and hide underlying implementation details, the memory layout of nodes is task-specific (a task can be an individual forward, backward, or aggregation phase) and sliced into frames.
A frame of a given node is a stack of consecutive resident memory, storing raw data and tensors.
To alleviate peak memory pressure, the memory can be allocated and released dynamically per frame on the fly.
More specifically, in the forward/backward phase, output tensors for each layer are calculated and released immediately after use.
As the allocation and de-allocation of tensors have a context-aware memory usage pattern, we design a tensor caching between frames and standard memory manipulation libraries to avoid frequently trapping into operating system kernel spaces. 

\textbf{\texttt{GraphView} and multi-versioned parameters.} 
To train multiple subgraphs concurrently, we design the abstraction \texttt{GraphView},
which maintains all key features of the underlying parallel graph storage including reused indexing, embedding lookup, and the distributed graph representation.
Implemented as a light-weighted logic view of the global graph, 
the \texttt{GraphView} exposes a set of interfaces necessary to all training strategies, 
and allows to conveniently communicate with storage.
Besides the global-, mini-, and cluster-batch, other training strategies can also be implemented based on \texttt{GraphView}.
And training tasks with \texttt{GraphView}s are scheduled in parallel.
That enables concurrently assigning the separated forward, backward, and aggregation phases to a training worker.
Due to the varied workloads of subgraphs, a work-stealing scheduling strategy is adopted to improve load balance and efficiency.

Figure~\ref{paralleltraining_fig} depicts the parallel batched training paradigm with graph view and multi-versioned parameter management.
In the figure, \texttt{ParameterManager} manages multiple versions of trainable parameters. 
In a training step, workers can fetch parameters of a specific version from \texttt{ParameterManager}, and use these parameters within the step.
For each worker, it computes on a local slice of the target subgraph being trained and uses a task queue (i.e., \texttt{TaskQueue}) to manage all the tasks assigned to itself and then execute them concurrently.
At the end of a training step, parameter gradients are aggregated and sent to \texttt{ParameterManager} for a version update.
\texttt{UpdateParam} performs the actual parameter update operations either in a synchronous or an asynchronous mode~\cite{ho2013more}.
It needs to be stressed that we will only use synchronous training for each test throughout this paper.

%
%
%
%
%
%

\section{Experiments and Results}

\subsection{Experiment Setups}
\label{sec:experiment_setups}
\textbf{Datasets Information.}
We evaluate the performance of our system by training node classification models using 7 datasets.
As shown in  Table~\ref{tbl:dataset}, the network sizes vary from small-, modest-, to large-scale. 
The first 6 datasets including Cora, Citeseer, Pubmed, Reddit, Amazon and ogbn-papers100M (Papers)~\cite{hu2020open} are publicly available and have only node attributes, without edge attributes or types.
Cora, Citeseer, and Pubmed~\cite{Citationdata} are three citation networks with nodes representing documents and edges indicating citation relationships between documents.
In the three datasets, the attribute of a node is a bag-of-words vector, which is sparse and high-dimensional, while the label of the node is the category of the corresponding document.
Reddit is a post-to-post graph in which one post forms one node and two posts are linked if they are both commented on by the same user.
In Reddit, a node label presents the community a post belongs to~\cite{hamilton2017inductive}.
Amazon is a co-purchasing graph, where nodes are products and two nodes are connected if purchased together.
In Amazon, node labels represent the categories of products~\cite{hu2021ogblsc}.
Papers is a directed citation network extracted from the Microsoft academic graph~\cite{wang2020microsoft} and consists of 0.1 billion nodes and 1.6 billion edges.
In this dataset, nodes denote papers and edges represent citation relationship.

The last Alipay dataset is our private industrial big graph, which contains 1.4 billion nodes and 4.1 billion attributed edges and is relatively sparse with a density of about 3.
In this dataset, nodes are users, attributes are user profiles and class labels are the financial risk levels of users.
Edges are built from a series of relations among users such as chatting, online financial cooperation, payment, and trade.
To the best of our knowledge, Alipay is the largest edge-attributed industry-scale graph ever used to test GNNs that use edge attributes in their graph convolutions in the literature.
In our experiments, we equally split the data set into two parts, one for training and the other for testing.

\begin{table}[t!]
\begin{center}
\caption{Information of the datasets used.}
\label{tbl:dataset}
\begin{tabular}{p{0.8cm}p{0.7cm}p{0.9cm}p{0.8cm}p{0.8cm}p{0.9cm}p{1.2cm}}
\toprule
Name &
  \#Nodes & \makecell{\#Node\\attr.} & \#Edges & \makecell{\#Edge\\attr.} & \makecell{Max\\cluster} & \makecell{\#clusters} \\
\midrule
Cora         & 2.7K         & 1.4K & 5.4K         & 0  &  $-$          & $-$ \\

Citeseer     & 3.3K         & 3.7K & 4.7K         & 0  &  $-$          & $-$  \\

Pubmed       & 19K        & 500  & 44K        & 0  &  $-$          & $-$  \\

Reddit       & 233K       & 602  & 11M  & 0  & 30K     & 4.3K \\

Amazon       & 2.4M     & 100  & 61M   & 0  & 36K   & 8.5K \\

Papers & 111M & 128 & 1.6B	& 0	& $-$ &$-$ \\

Alipay       & 1.40B   & 575  & 4.14B & 57 &  24M & 4.9M  \\
\bottomrule
\end{tabular}
\end{center}
\vspace{-0.2cm}
\end{table}

\textbf{GNN Training Settings.}
\label{sec:training_settings}
We employ the node classification task as the application and use the popular GCN~\cite{kipf2016semi} model and our in-house developed GAT-E model for performance comparison.
Herein, we would like to emphasize that in these tests, our purpose is not to conclude which DL framework or GNN implementation is superior to others, but to demonstrate that \projectname is capable of learning GNNs as well as existing frameworks (Section~\ref{appendix:gat} in supplementary material also gives another showcase with GAT~\cite{2017GAT} model).
Note that although there are only node classification tasks in our tests, our system can support other types of tasks with moderate changes, such as revising the decoders to accommodate specific tasks and keeping the graph embedding encoder part unchanged.

Different datasets are configured to have varied hidden layer sizes. Specifically, hidden layer sizes are 16, 128, and 200 for the three citation networks (i.e., Cora, Citeseer, and PubMed), Reddit, and Amazon.
Except for Amazon, all the others have their sub-sets for validation.
Moreover, the latter enables dropout for each layer, whereas the former does not.
In terms of global-batch, we train the model up to 1,000 epochs and select the trained model with the highest validation accuracy to test generalization performance for the latter.
With respect to mini-batch, early stop is exerted once stable convergence has reached for each dataset.
In all tests, cross-entropy loss is used to measure convergence, and prediction results are produced by feeding node embeddings to a Softmax layer.

For global-batch, we set 500 epochs for Reddit at maximum, and activate early stop as long as validation accuracy reaches stable. Meanwhile, a maximum number of 750 epochs is used for Amazon.
For mini-batch, each training step randomly chooses 1\% labeled nodes to form the initial batch for Reddit, and 0.1\% for Amazon.
This results in a batch size of 1,500 for the former and 1,710 for the latter.
We train Reddit for 600 steps and Amazon for 2,750 steps with respect to mini-batch.
Note that as Reddit (and Amazon) is a dense co-comment (and co-purchasing) network, the two-hop neighbors of only 1\% (and 0.1\%) labeled nodes almost touch 80\% (and 65\%) of all nodes.
For cluster-batch, each training step randomly chooses 1\% clusters to form the initial batch for Reddit and Amazon, where clusters are created beforehand.
In addition, cluster-batch applies the same early stop strategy as mini-batch.

\subsection{Accuracy Assessment}
\subsubsection{Evaluation on Public Datasets.}
On the public datasets, we train a two-layer GCN model~\cite{kipf2016semi} using our system and compare the performance with some well-known counterparts, including a TensorFlow-based implementation (TF-GCN) from~\cite{kipf2016semi}, a DGL-based one (DGL), Cluster-GCN and 3 sampling-based ones: VR-GCN, GraphSAGE~\cite{hamilton2017inductive}, and GraphSAINT~\cite{zeng2019graphsaint}.
We exclude FastGCN from our comparisons because it is outperformed by Cluster-GCN and GraphSAINT as shown in ~\cite{zeng2019graphsaint, ClusterGCN}.
Note that our system is designed for non-sampling graph learning, and our performance evaluation does not intend to compete in terms of state-of-the-art accuracy on these public datasets, but to demonstrate that our system can train as good GNN models as existing works.

\textbf{Comparison with non-sampling-based methods.}
\label{sec:non-sampling}
First, we compare our system with TF-GCN, DGL, and Cluster-GCN to demonstrate that our system can achieve highly competitive or superior generalization performance on the same datasets and models, even without using sampling.
Table~\ref{tbl:result_small} shows the performance comparison on the three small-scale citation networks, i.e., Cora, Citeseer, and PubMed~\cite{Citationdata}.
For DGL and Cluster-GCN, since they did not completely expose their hyper-parameters, this poses challenges for result reproduction.
Both of them have been re-evaluated with the hidden layer size set to 16 for each dataset as described in Section~\ref{sec:training_settings}.
It is observed that DGL almost reproduces its original result in~ \cite{wang2019deep} for each dataset, but Cluster-GCN yields far inferior results to those in ~\cite{ClusterGCN} for Cora and Pubmed (note that Citeseer is not used in ~\cite{ClusterGCN}).
Both \projectname and TF-GCN use the same set of hyper-parameter values, including learning rate, dropout keep probability, regularization coefficient, and batch size, as proposed in~\cite{kipf2016semi}, and train the model 300 steps for mini-batch.
From Table~\ref{tbl:result_small}, global-batch yields the best accuracy for each dataset, with the exception that its performance is neck-by-neck with that of mini-batch on Citeseer.
Mini-batch also outperforms DGL, TF-GCN and Cluster-GCN for each case.

%
%
\begin{table}[t!]
\begin{minipage}{\linewidth}
\centering
\caption{Comparison to counterparts without sampling.}
\label{tbl:result_small}
\begin{tabular}{lccccc}
\toprule
					& \multicolumn{4}{c}{Accuracy in Test Set (\%)}                    \\ \cline{2-6} 
\multirow{-2}{*}{Dataset} & \makecell{GCN \\ w/ GB}
                          & \makecell{GCN \\ w/ MB}
                          & \makecell {GCN \\ On DGL}
                          & \makecell{GCN \\ On TF}
                           & \makecell{Cluster\\-GCN} \\
\midrule
 Cora         & \textbf{82.70} & 82.40 & 80.50 & 81.50 & 70.47 \\
 Citeseer     & \textbf{71.90} & \textbf{71.90} & 70.36 & 70.30 & 59.43 \\
 Pubmed       & \textbf{80.00} & 79.50 & 79.06 & 79.00 & 75.14 \\
 \bottomrule
\multicolumn{6}{l}{GB: global-batch, MB: mini-batch, and CB: cluster-batch}
\end{tabular}
\end{minipage}
\begin{minipage}{\linewidth}
\centering
\caption{Comparison to counterparts with sampling.}
\label{tbl:result_middle}
\begin{tabular}{p{0.75cm}p{0.44cm}p{0.44cm}p{0.44cm}p{0.7cm}ccc}
\toprule
            	& \multicolumn{7}{c}{Accuracy in Test Set (\%)}                    \\ \cline{2-8} 
\multirow{-2}{*}{Dataset} & GB & MB & CB & \makecell{VR-\\GCN} & \makecell{Cluster\\-GCN} &\makecell{Graph\\SAGE} & \makecell{Graph\\SAINT} \\
\midrule
Reddit         & \textbf{96.44}  & 95.84  &95.60 & 62.48 & 96.23 & 96.20 & \textbf{96.44} \\
Amazon         & \textbf{89.77}  & 87.99  &88.34  & 71.77 & 75.66 & 77.13 & 76.38 \\
\bottomrule
\end{tabular}
\end{minipage}
\begin{minipage}{\linewidth}
\begin{center}
\caption{Accuracy comparison on Alipay dataset.}
\label{tbl:result_large}
\begin{tabular}{llll}
\toprule
                                      & \multicolumn{2}{r}{Performance in Test Set (\%)}                    \\ \cline{2-3} 
\multirow{-2}{*}{Strategies} & F1 Score                      & AUC & \multirow{-2}{*}{Time (h)}                     \\
\midrule
\begin{tabular}[l]{@{}r@{}} Global-batch\end{tabular}         & 12.18 & 87.64 & 30 \\

Mini-batch                                                                    & 13.33 & 88.12 & 36 \\

Cluster-batch                                                               & \textbf{13.51} & \textbf{88.36}& \textbf{26} \\
\bottomrule
\end{tabular}
\end{center}
\end{minipage}
\vspace{-0.2cm} 
\end{table}
\textbf{Comparison with sampling-based methods.}
\label{sec:sampling}
Second, we compare our implementations (without sampling) with VR-GCN, GraphSAGE, GraphSAIT, and Cluster-GCN on the two modest-scale datasets: Reddit and Amazon.
VR-GCN adopts variance reduction, GraphSAGE uses random sampling, GraphSAINT introduces subgraph sampling, whereas Cluster-GCN does not use sampling same as \projectname.
All of these GNNs train models in a mini-batched manner. Similar to the above non-sampling-based comparisons, we re-evaluate them rather than directly use the results from their corresponding publications, where the hidden layer size is set to 128 for Reddit and 200 for Amazon as stated in Section~\ref{sec:training_settings}.
It needs to be stressed that since GraphSAINT offers three random samplers including random node samplers, random edge samplers, and random walk samplers, we have assessed all of the samplers and singled out the best to report for each dataset.

Table~\ref{tbl:result_middle} gives the test accuracy of each experiment.
On both datasets, global-batch yields the best accuracy, while mini-batch performs the worst with cluster-batch in between.
Moreover, VR-GCN shows considerably worse performance than any of the counterparts on each dataset.
In terms of Reddit, GraphSAINT also performs best same with our global-batch, while Cluster-GCN outperforms GraphSAGE by a tiny margin.
For Amazon, however, GraphSAGE becomes better than Cluster-GCN.
Meanwhile, GraphSAINT is still superior to VR-GCN, Cluster-GCN, and GraphSAGE.
It is worth mentioning that the relatively lower performance of our cluster-batch may be caused by multiple factors, e.g. number of clusters,  cluster batching,  and the quality of graph partitioning.
In sum, based on the aforementioned observations, it is reasonable to draw the conclusion that sampling-based training methods are not always better than non-sampling-based ones.
\subsubsection{Evaluation on Alipay dataset}
\label{sub_section:alipay}
In this test, we use our in-house developed GAT-E model, a graph attention network that incorporates edge attributes to attention computation along with node attributes.
GAT-E is a simplified version of the GIPA~\cite{zheng2021gipa} algorithm and is trained on the Alipay dataset with all of the three training strategies.

Table~\ref{tbl:result_large} shows the accuracy on the Alipay dataset.
We run 400 epochs for global-batch, and 3,000 steps for all the others.
Cluster-batch performs the best, while mini-batch outperforms global-batch, in terms of both F1 score and accuracy.
In terms of convergence speed in the same distributed environment, we can observe that cluster-batch converges the fastest. The second fastest method is global-batch, and the third one is mini-batch. 
Specifically, the overall training time of 1,024 workers is 30 hours for global-batch, 36 hours for mini-batch, and 26 hours for cluster-batch, and the peak memory footprint per worker is 12~GB, 5~GB, and 6~GB respectively.
\subsection{Scalability Assessment}
\subsubsection{Comparison of Different Training Strategies}
\label{sec:alipay_scale}
\begin{figure*}[t!]
\centering
\subfigure[]{
\begin{minipage}[b]{0.3\linewidth}
\includegraphics[width=\linewidth]{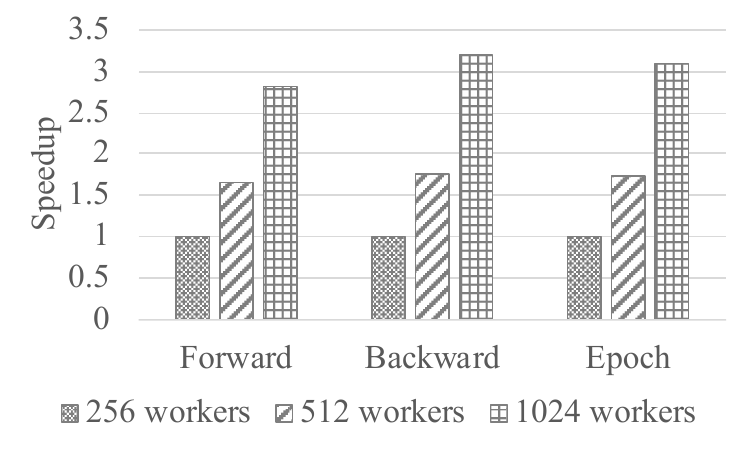}
\label{fig:gb_scale}
\vspace{-0.7cm}
\end{minipage}
}
\subfigure[]{
\begin{minipage}[b]{0.3\linewidth}
\includegraphics[width=\linewidth]{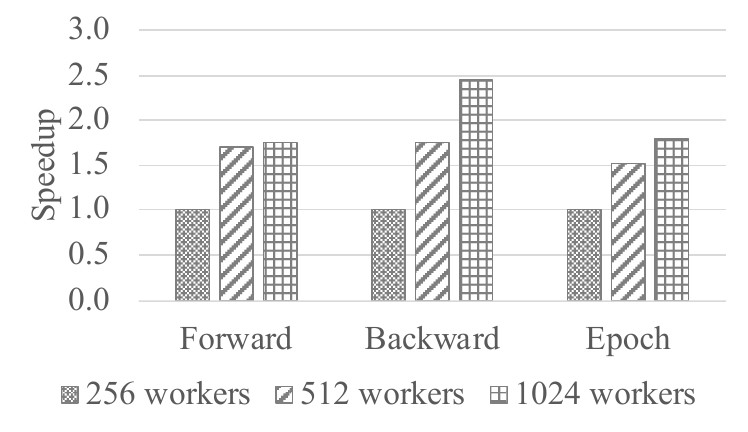}
\vspace{-0.7cm}
\label{fig:cb_scale}
\end{minipage}
}
\subfigure[]{
\begin{minipage}[b]{0.3\linewidth}
\includegraphics[width=\linewidth]{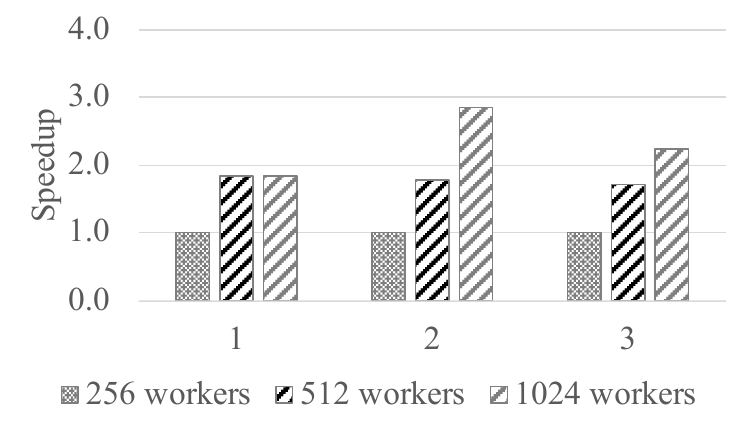}
\vspace{-0.7cm}
\label{fig:mb_scale}
\end{minipage}
}
\vspace{-0.3cm}
\caption{Scalability of \projectname on Alipay for (a) global-batch, (b) cluster-batch and (c) mini-batch.}
\label{fig:scales}
\vspace{-0.6cm}
\end{figure*}
\begin{figure*}[t!]
\subfigure[] {
\begin{minipage}[b]{0.3\linewidth}
\centering
\includegraphics[width=\linewidth]{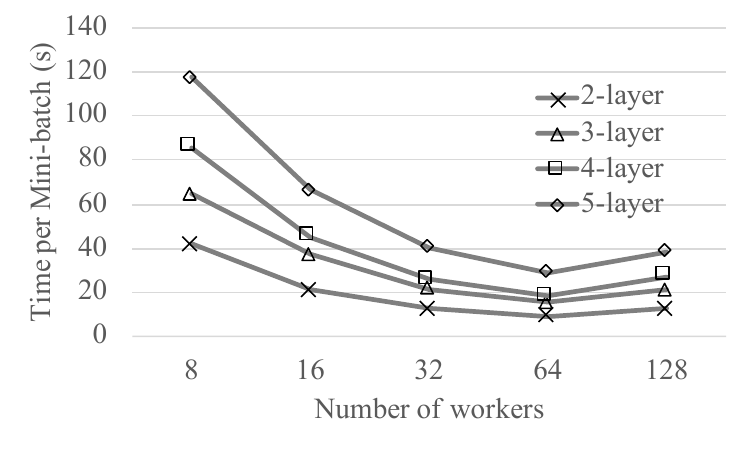}
\vspace{-0.7cm}
\label{fig:distdgl_scale}
\end{minipage}
}
\subfigure[] {
\begin{minipage}[b]{0.3\linewidth}
\centering
\includegraphics[width=\linewidth]{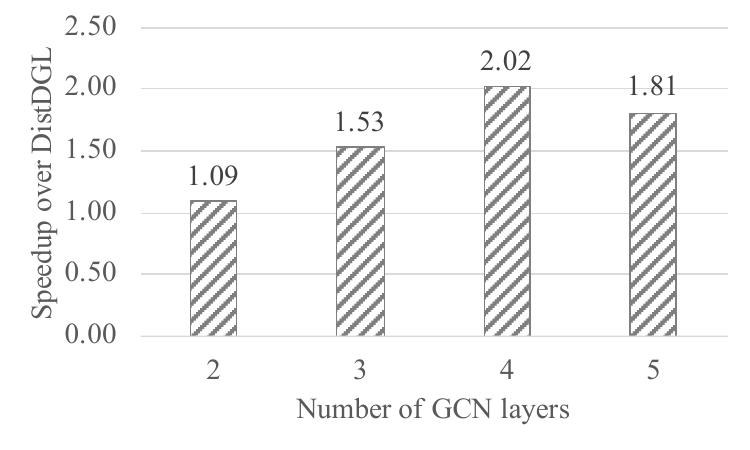}
\vspace{-0.7cm}
\label{fig:distdgl_speedups}
\end{minipage}
}
\subfigure[] {
\begin{minipage}[b]{0.3\linewidth}
\centering
\includegraphics[width=\linewidth]{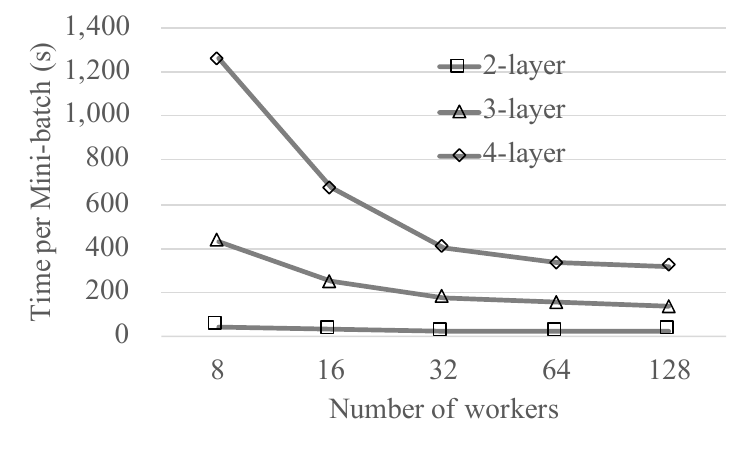}
\vspace{-0.7cm}
\label{fig:graphtheta_papers100m_scale}
\end{minipage}
}
\vspace{-0.3cm}
\caption{(a) our scalibility on Reddit; (b) our speedups over DistDGL on Reddit; and (c) our scalability on Papers.}
\label{fig:scalability_speedups}
\vspace{-0.6cm}
\end{figure*}

We evaluate the capability of strong scaling to big graphs of \projectname with respect to the number of workers using Alipay dataset.
Each worker is equipped with one computing thread and runs in a Linux CPU Docker.
Due to the large memory footprint of Alipay dataset, we start with 256 workers and use this performance as the baseline.

Figures~\ref{fig:gb_scale}, \ref{fig:cb_scale} and \ref{fig:mb_scale} illustrate the speedup results in the function of the number of workers for global-batch, cluster-batch, and mini-batch respectively.
From the figures, we can see that each of the three training strategies can scale to 1,024 workers, and the speedups for the forward, backward, and full training steps are consistent in terms of the scaling size for each training strategy.
This can be explained by our observation that the neural network functions are usually compute-intensive and thus lead to better computation and communication overlap, significantly lowering the impact of communication.
By investigating the speedups of the three training strategies, it can be observed that global-batch has the best scalability, followed by cluster-batch and mini-batch in decreasing order.
This is because ($i$) global-batch has relatively balanced workloads among workers, since all nodes in the graph participate in the computation simultaneously, and ($ii$) cluster-batch has better data locality among distributed machines, resulting in less inter-machine communication than mini-batch.

In the following, we will analyze the speedups and parallel efficiency gained as the number of workers varies.
For global-batch, by increasing the number of workers from 256 to 512, the forward runs $1.66\times$ faster, the backward $1.75\times$ faster, and the full training step $1.72\times$ faster.
Furthermore, when increasing the number of workers further to 1,024, the speedups become $2.81$, $3.21$, and $3.09$ times respectively.
In terms of parallel efficiency, the forward achieves $83\%$ (and $70\%$),  backward $87\%$ (and $80\%$), and full training steps $86\%$ (and $77\%$) by using 512 workers (and 1,024 workers), respectively.

In terms of cluster-batch, when moving from 256 to 512 (and 1024) workers, the speedup is 1.70 (and 1.75) for the forward, 1.74 (and 2.45) for the backward, and 1.52 (and 1.80) for the full training step.
In this case, the corresponding parallel efficiency becomes $85\%$ (and $44\%$) for the forward, $87\%$ (and $61\%$) for the backward, and $76\%$ (and $45\%$) for the full training step.

With respect to mini-batch, compared to the baseline performance at 256 workers, the speedup and parallel efficiency at 512 workers is 1.83 and $91\%$ for the forward, 1.77 and $88\%$ for the backward, and 1.71 and $86\%$ for the full training step, respectively.
Meanwhile, the corresponding values at 1024 workers are 1.84 and $46\%$ for the forward, 2.85 and $71\%$ for the backward, and 2.23 and $56\%$ for the full training step, respectively.
\subsubsection{Comparison with DistDGL}
\label{sec:comp_distdgl}
In this section, we will compare with DistDGL~\cite{zheng2020distdgl, zheng2022distributed} in terms of scalability, as well as the best performance with the same computing resources, by training the GCN model on the Reddit dataset.
In this test, node sampling in the node data loader of DistDGL is disabled for the sake of fair comparison.
Both \projectname and DistDGL are launched to synchronously train four GCN models of different layers ranging from 2 to 5, with mini-batch on a cluster of 8 virtual machines with 64 CPU cores and 300 GB RAM each.

For scalability assessment, we set to use 4 cores for each worker/trainer and keep the overall batch size of value 24K unchanged (i.e. the batch size per worker/trainer times the number of workers/trainers is equal to 24K).
Considering that DistDGL runs a distributed graph storage server within each machine, the computational resources of each machine have to be split among the servers and all trainers running in the same machine.
Herein, we set the number of threads per server to be $\max \{16, 64 - 4\times p\}$, where $p$ represents the number of trainers per machine.

Figure~\ref{fig:distdgl_scale} shows the scalability of \projectname, which demonstrates consistent scalability for all models.
Specifically, each model scales to 64 workers with a slight performance degradation at 128 workers.
Section~\ref{appendix:sec_distgdl_scale} in supplementary material shows that DistDGL does not scale at all for each model.
We analyze that this phenomenon is caused by the increasing amount of redundant computation among batches as the number of trainers grows. 
Specifically, as more trainers are launched, the batch size per trainer is reduced proportionally since the overall batch size is invariant in our synchronous training.
In this case, some neighbor nodes shared by the input target nodes of a large batch have to be replicated within the set of small batches.
These replicated nodes will be computed multiple times between batches and thus lower training speed and scalability.
However, \projectname does not suffer from this problem, because the subgraph constructed from the target nodes is independent of the number of workers and its overall amount of computation does not grow as more workers join.
In addition, our hybrid-parallel execution can further reduce runtime by using more workers.

Besides scalability, we also compare with the best performance of DistDGL gained on this cluster of virtual machines.
As suggested by one DistDGL developer, we launch only one trainer in each machine but tune the number of threads assigned to the trainer and the distributed server. Refer to Section~\ref{appendix:sec_distdgl_tune} in the supplementary material for more details.
Figure~\ref{fig:distdgl_speedups} illustrates the speedup of \projectname over DistDGL for each model.
More specifically, the speedup is 1.09, 1.53, 2.02, and 1.81 for the 2-layered, 3-layered, 4-layered, and 5-layered model, respectively.
\subsubsection{Comparison with GraphLearn}
\label{sec:comp_graphlearn}
In this section, we will compare with GraphLearn in terms of scalability on the same CPU cluster as in Section~\ref{sec:comp_distdgl}.
We use four GCNs of the different number of layers (i.e. 2, 3, and 4) on the Reddit and Papers datasets, but exclude the 5-layered GCN, because GraphLearn does not work on it even with a small sampling setting to be described below.
Our synchronous distributed training keeps the overall batch size of 24K for Reddit and 12K for Papers, and sets hidden layer sizes to 128.
Same as in Section~\ref{sec:comp_distdgl}, GraphTheta still configures 4 cores for each worker.
Figure~\ref{fig:graphtheta_papers100m_scale} illustrates the scalability of \projectname on Papers, where our performance gets better as the number $w$ of workers increases for the 3- and 4-layered GCNs, but encounters marginal degradation from 64 to 128 workers for the 2-layered one.

GraphLearn provides multiple sampling strategies, among which the "full" sampling strategy is chosen.
This strategy returns all neighbors if the number of neighbors for a hop is less than or equal to the threshold $nbr\_num$ and otherwise truncates by $nbr\_num$.
We have attempted to disable sampling for GraphLearn by setting $nbr\_num$ to a large value under the "full" strategy but encountered socket errors.
Hence, we keep sampling enabled and employ two sampling settings concerning $nbr\_num$.
One configures $nbr\_num$ to relatively small values 10,5,3,3 (following ByteGNN~\cite{zheng2022bytegnn}), which correspond to hops from 1 to 4, while the other sets $nbr\_num$ to relatively large values 20,10,10,2.

In GraphLearn, each graph server creates a thread pool of 32 threads by default to process concurrent sampling queries.
We have tried to reduce the thread pool size to some small values but encountered errors.
Therefore, we keep using 32 concurrent thread settings throughout our tests.
Table~\ref{tab:graphlearn} gives the performance with $w$ varying from 8 to 32.
It needs to be stressed that launching $w>32$ workers will result in socket errors. 
From the table, for every combination of the dataset and sampling setting, it is surprising to observe super-linear speedups along with a doubling increase of $w$, with the performance of 8 workers as the baseline.
After communicating with a GraphLearn developer, we analyze that this phenomenon is due to the following two reasons.
One is that the thread pool can run 32 concurrent threads at maximum.
Therefore, the processing capability of concurrent queries could ideally increase at the same rate as $w$ grows.
The other is that multiple workers concurrently run on the same machine and thereby can benefit from faster intra-machine communication than inter-machine one.

\begin{table}[t!]
\centering
\caption{Average runtimes per mini-batch for GraphLearn.}
\label{tab:graphlearn}
\begin{tabular}{ccccccc}
\toprule
& \multicolumn{3}{c}{Reddit (\#workers)} & \multicolumn{3}{c}{Papers (\#workers)} \\
\cline{2-7}
\multirow{-3}{*}{GCNs} & 8 & 16 & 32 & 8 & 16 & 32\\
\midrule
\multicolumn{7}{c}{Sampling setting 10,5,3,3 (time in seconds)} \\
\cline{2-7}
2-layer &	13.6 &	4.4&	1.6&	3.6 & 1.2 & 0.5 \\
3-layer	&614.8 &	153.6&	40.8&	109.8& 30.0 & 8.5\\
4-layer &8863.5 & 2301.1 & 565.6& 1759.4 & 433.8 & 112.8\\
\midrule
\multicolumn{7}{c}{Sampling setting 25,10,10,2 (time in seconds)} \\
\cline{2-7}
2-layer	& 32.2 & 11.1 & 5.7 & 9.9 & 3.7 & 1.8\\
3-layer & $-$ &  $-$ & $-$ & 984.7 & 253.2 & 72.3\\
4-layer & $-$ & $-$ & $-$ & $-$ & $-$ & $-$ \\
\bottomrule
\multicolumn{7}{l}{$-$ indicates the results are unavailable caused by socket errors.}
\end{tabular}
\vspace{-0.2cm} 
\end{table}

Even though sampling reduces the computational overhead of GraphLearn and thus results in unfair comparisons with sampling-free systems, we still compare with GraphLearn in terms of the best performance on the same machines as done in Section~\ref{sec:comp_distdgl}.
Our evaluation shows that on Reddit, \projectname yields a speedup of 2.61 (and 30.56) for the 3-layered (and 4-layered) GCN.
As for Papers, \projectname does not demonstrate speed advantages to GraphLearn, as expected, but the latter completely failed in the 4-layered GCN when using sampling setting 25,10,10,2.
In this sense, \projectname is still superior.
Finally, note that GraphLearn uses a Python user-defined function (UDF) to create sparse tensors for a mini-batch.
Hence, we expect that its performance could get further improved if the UDF was programmed in some more efficient language.
\begin{figure*}[t!]
\centering
\subfigure[Global-batch]{
\begin{minipage}[b]{0.3\linewidth}
\includegraphics[width=\linewidth]{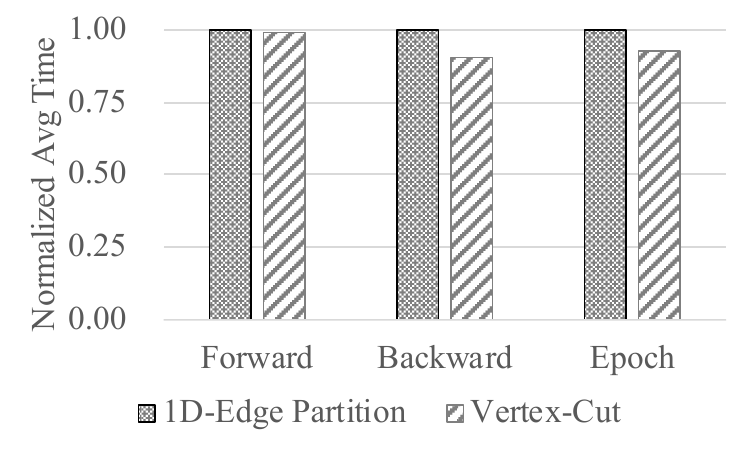}
\vspace{-0.6cm}
\label{fig:partition_gb_amazon}
\end{minipage}
}
\subfigure[Cluster-batch]{
\begin{minipage}[b]{0.3\linewidth}
\includegraphics[width=\linewidth]{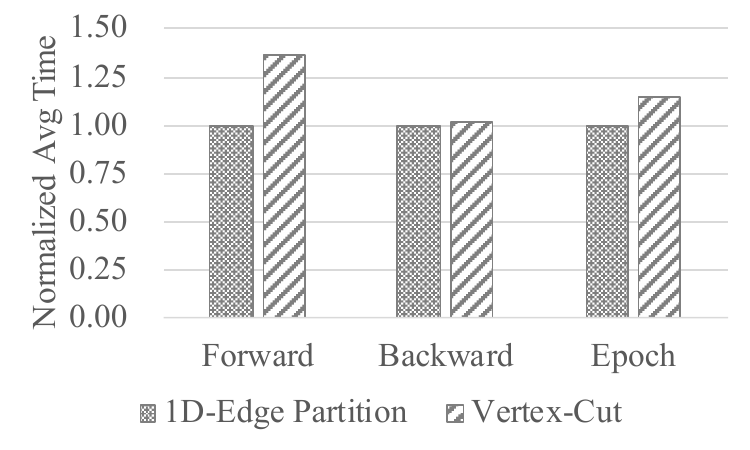}
\vspace{-0.6cm}
\label{fig:partition_cb_amazon}
\end{minipage}
}
\subfigure[Mini-batch]{
\begin{minipage}[b]{0.3\linewidth}
\includegraphics[width=\linewidth]{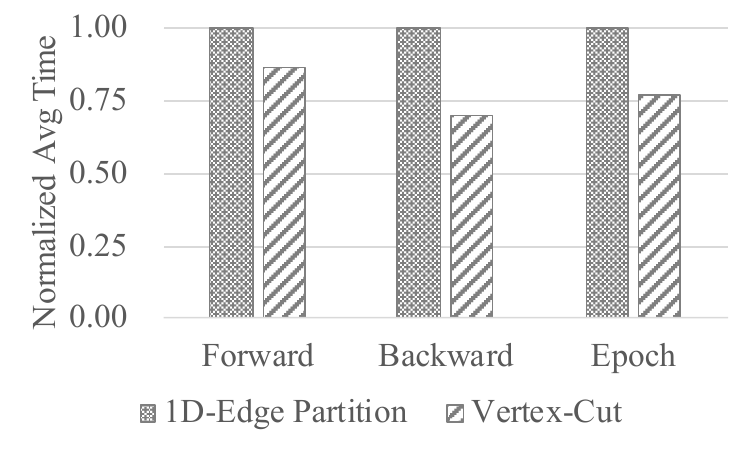}
\vspace{-0.6cm}
\label{fig:partition_mb_amazon}
\end{minipage}
}
\vspace{-0.3cm}
\caption{Comparison between vertex-cut and 1D-edge partitioning methods on Amazon for each training strategy.}
\label{fig:partition_amazon}
\vspace{-0.6cm}
\end{figure*}
\subsection{Effect of Graph Partitioning Methods}
\label{result:partition}
In this section, we will evaluate the effect of the following two graph partitioning methods: \texttt{vertex-cut} and \texttt{1D-edge partition}, on the execution of different training strategies using the Amazon dataset.

\texttt{vertex-cut} is a 2D-grid partitioning method, which targets to evenly distribute edges to all workers.
Given an edge, \texttt{vertex-cut} determines the worker it is assigned to by computing a hash value from the source and destination nodes of the edge.
\texttt{1D-edge partition} aims to evenly distribute source (or destination) nodes to all workers.
Given an edge, \texttt{1D-edge partition} determines its destination worker by computing a hash value from the source (or destination) node.
In our implementation, we compute the hash value from the source node, which is also the master node, but allow for users to configure to use the destination.
\texttt{vertex-cut} has the advantage of addressing the highly skewed node degree distribution problem, while \texttt{1D-edge partition} leads to better edge locality for the perspective of source (or destination) nodes.

In \projectname, we set \texttt{1D-edge partition} as the default partitioning method, based on the following two considerations.
On one hand,  both the use of edge attributes and the computation of edge attention and embedding has been becoming more and more popular in GNNs~\cite{wu2020comprehensive,zhang2020deep,chaudhari2021attentive}.
With \texttt{vertex-cut}, the edges of a given master node are likely distributed among multiple workers.
In this case, both the loading of edge attributes and the computation of edge attention and embedding will incur considerable extra communication. 
On the contrary, \texttt{1D-edge partition} always places a master node and all of its edges together in the same worker.
This way, we can load the edge attributes and compute the edge attention and embedding for the master node locally, with no need of extra communication.
On the other hand, in practice, we observe that \texttt{vertex-cut} usually has a higher peak memory footprint than \texttt{1D-edge partition}.
Taking the benchmark on Amazon as an example, the peak memory of \texttt{vertex-cut} is about 20\% larger than that of \texttt{1D-edge partition} on average.
Nevertheless, users can still specify their preferred partitioning method at runtime.

Figures~\ref{fig:partition_gb_amazon}, \ref{fig:partition_cb_amazon} and \ref{fig:partition_mb_amazon} show the normalized average runtimes of the forward, the backward and the full epoch steps, respectively in terms of global-batch,  cluster-batch, and mini-batch.
The normalization uses the corresponding runtime of \texttt{1D-edge partition} as the baseline.
From the figures, \texttt{vertex-cut} is superior to \texttt{1D-edge partition} in terms of global-batch and mini-batch, and inferior to the latter with respect to cluster-batch.
Based on these observations, we would like to make the following suggestion. If there is sufficient memory in distributed machines, we recommend users try \texttt{vertex-cut} first in an optimistic manner. Otherwise, \texttt{1D-edge partition} is supposed to be in preference.
%

\section{Related Work}
\label{section:related_work}
Existing GL frameworks for GNNs employ either shared-memory systems~\cite{wang2019deep,eksombatchai2018pixie,ying2018graph,lin2020pagraph} or distributed computing systems~\cite{zheng2020distdgl,zheng2022distributed,ying2018graph,wan2022pipegcn,wang2021flexgraph,gandhi2021p3,thorpe2021dorylus}. Most are designed for mini-batched training~\cite{zhu2019aligraph,zheng2022bytegnn,zhang2020agl}, while some others are for global-batch~\cite{liao2018graph,ma2019neugraph,jia2020improving,ying2018graph,wan2022pipegcn}.
Herein, we will mainly analyze the pros and cons of those works that are capable of processing billion-scale networks.

PinSage~\cite{ying2018graph} trains GNNs with sampling enabled and uses MapReduce to implement non-sampling GNN inference.
However, MapReduce incurs considerable manual GNN translation overhead.
Specifically, users have to re-write and translate GNN models into MapReduce procedures and make endeavors to guarantee the correctness of this translation.
AGL~\cite{zhang2020agl} extends the idea of PinSage by additionally employing MapReduce to extract training subgraphs offline and save them on disk for future use, while still using MapReduce to implement inference.
AGL can consume a huge amount of disk to store subgraphs and still undergoes GNN translation burden with respect to inference.

AliGraph~\cite{zhu2019aligraph} introduces distributed graph servers to perform sampling.
Compared to MapReduce-based methods, AliGraph does not require separating training from inference, thus no need of GNN translation, but demands sampling for inference.
One good thing is that AliGraph generates subgraphs at runtime and discards them once they have been consumed, thus not demanding extra persistent storage of these subgraphs.
Nonetheless, AliGraph does not always work well because DL workers have to pull all necessary neighbor information from remote servers and suffer from the inference instability problem caused by sampling.
ByteGNN~\cite{zheng2022bytegnn} optimizes AliGraph on CPUs to improve CPU utilization and reduce network communication overhead caused by sampling.
DistDGL~\cite{zheng2020distdgl, zheng2022distributed} extends DGL~\cite{wang2019deep} and adopts a similar idea to AliGraph, but deploys graph servers as a separate service independent of DL workers, while the latter binds graph servers to DL workers and run them in the same processes.
This separate deployment makes DistDGL clients have to always pull subgraphs from servers remotely.
P3~\cite{gandhi2021p3} is also implemented based on DGL by applying model-parallel only to the first GCN layer but data-parallel to the rest.
This way the memory problem with big graphs and deep GNNs can only be solved in the first layer but remains for the rest.
DistGNN~\cite{md2021distgnn} intends to optimize DGL from the perspectives of shared memory implementation and communication reduction and avoidance but is only oriented to distributed global-batched training.

Dorylus~\cite{thorpe2021dorylus} adopts serverless computing~\cite{fouladi2017encoding, klimovic2018pocket, jonas2017occupy} and separates graph operations from the execution of neural network functions.
However, it needs to load all needed neighbors of given target nodes into shared memory.
FlexGraph~\cite{wang2021flexgraph} implements graph servers with GRAPE~\cite{fan2017grape} and builds a hierarchical dependency graph to facilitate mini-batched neighbor aggregation.
However, this does not address the inherent limitation embedded in mini-batch.
GNNLab~\cite{yang2022gnnlab} targets single-machine multi-GPU training and dedicates GPUs to whether sampling or model training.
Unlike the above works almost all targeting mini-batch,
PipeGCN~\cite{wan2022pipegcn} introduces a pipeline feature communication mechanism to optimize global-batch.
In addition, some works enhance GL from the perspectives of FPGA~\cite{auten2020hardware,li2022hyperscale}, distributed file system~\cite{van2022gdll}, hybrid data dependency~\cite{wang2022neutronstar}, approximate PageRank sampling~\cite{bojchevski2020scaling}, data efficiency~\cite{zhang2021grain} and federated learning~\cite{wang2022federatedscope,wu2022federated,pei2021decentralized,he2021fedgraphnn,wu2021fedgnn,guan2021federated,zhou2020vertically,jiang2022federated,mei2019sgnn}.

Our system intrinsically differs from existing methods in the following aspects. First, our system is built from scratch based on distributed graph processing engine with neural network operators as UDFs and introduces an NN-TGAR abstraction to bridge the gap between conventional graph processing and deep learning. Second, our system employs a hybrid-parallel execution, which computes each batch by a group of workers (processes) in a distributed manner, and flexibly addresses the challenges imposed by the combinatorial optimizations of several factors in GNN learning, including large or highly-skewed graphs, multiple training strategies, deep GNNs and free from neighbor sampling. Finally, it is worth noting that although general-purpose graph processing systems~\cite{malewicz2010pregel, liu2011parallelized, gonzalez2014graphx, gonzalez2012powergraph, wang2016gunrock,zhu2016gemini, chen2019powerlyra, huan2022tegraph,heng2022bring} are incapable of directly supporting GNN learning, their graph processing techniques can be taken as the infrastructure of graph learning systems.

\vspace{-0.2cm}
\section{Conclusion}
In this paper, we have presented \projectname, a distributed and scalable GNN learning system, which is implemented based on a graph processing programming model and introduces an NN-TGAR abstraction to ease graph convolution implementations.
Differing from conventional data-parallel and model-parallel paradigms, \projectname adopts a hybrid parallel execution, which distributively computes each batch of graph data by a group of processes/workers. Moreover, our system supports flexible training strategies and performs inference through a unified implementation with training.
These features distinguish our system from existing frameworks.

Extensive experiments show that \projectname can scale to 1,024 workers in distributed CPU virtual machines.
Considering the affordable access to CPU virtual machine clusters in public clouds like Aliyun, Amazon Web Services, Google Cloud, and Azure, our system shows promising potential to achieve low-cost graph learning on big graphs.
Moreover, our system demonstrates comparable or better generalization than the well-known GCN implementations on a diverse set of networks.
Compared with DistDGL, \projectname yields much better scalability and runs up to $2.02\times$ faster in terms of the best performance on the Reddit dataset using the same machines.
On the same dataset, meanwhile, \projectname can run up to $2.61\times$ (and $30.56\times$) faster than GraphLearn when training the 3-layered (and 4-layered) GCN.
In particular, \projectname can scale well with good parallel efficiency on the large edge-attributed Alipay dataset of 1.4 billion nodes and 4.1 billion edges.
Meanwhile, cluster-batch is observed capable of obtaining the best generalization and the fastest convergence speed on Alipay.
This suggests that it is worthwhile of developing new graph learning systems to allow for exploring a diversity of training strategies to meet the different requirements of various applications.

It is worth mentioning that the internal code name of \projectname is \interalname (\underline{G}raph \underline{E}xtended and \underline{A}ccelerated \underline{Learning}) and it has been deployed in production and used by several businesses in mobile Alipay App including Zhima Credit and risk management.
Currently, \projectname merely runs on CPUs and the performance of its implementation based on fine-grained vertex programs still has a large room for improvement.
Therefore, the performance improvement of \projectname in favor of GPUs is part of our future work.
Nonetheless, through this system, we have proven the feasibility of developing a distributed graph learning system based on graph processing.
This characteristic enables our system to execute graph processing and learning procedures in the same program, thereby opening up more opportunities to address graph intelligence problems involving diverse graph processing techniques (e.g. graph mining~\cite{teixeira2015distributed,talukder2016distributed,chen2020pangolin}).

\begin{acks}
We thank Kefeng Deng and other members of our GeaLearn team in Ant Group, including Wei Qin, Zhiqiang Guo, Yice Luo, Peng Du, Yue Jin and Xiabao Wu, for their contributions to this project.
We would like to thank Yanminng Fang for building the Alipay dataset. Here is the Data Protection Statement:
1. The data used in this research does not involve any Personal Identifiable Information(PII).
2. The data used in this research were all processed by data abstraction and data encryption, and the researchers were unable to restore the original data.
3. Sufficient data protection was carried out during the process of experiments to prevent the data leakage and the data was destroyed after the experiments were finished.
4. The data is only used for academic research and sampled from the original data, therefore it does not represent any real business situation in Ant Financial Services Group.

\end{acks}


\bibliographystyle{ACM-Reference-Format}
\bibliography{sample}

\clearpage
\begin{appendices}
\setcounter{table}{0}
\setcounter{figure}{0}
\renewcommand{\thetable}{A\arabic{table}}
\renewcommand{\thefigure}{A\arabic{figure}}
\section{Theoretical Analysis}
\subsection{Equivalence Relation}
\label{equivalence_relation}

Indeed, the propagation on graphs and sparse matrix multiplication is equivalent in the forward.
The general convolutional operation on graph $\mathcal{G}$ can be defined \cite{kipf2016semi}\cite{NIPS2016_6081} as, 
\begin{equation}
\bm{x}\,*_{\mathcal{G}}\,\mathcal{K}=\bm{U}((\bm{U}^{T}\bm{x})\odot(\bm{U}^{T}\mathcal{K})), 
\label{eq:kernel}
\end{equation}
where $U=[u_{0},u_{1},...,u_{N-1}]\in\mathbb{R^{\mathrm{N\times N}}}$ is the complete set of orthonormal eigenvectors of normalized graph Laplacian $\bm{L}$,
$\odot$ is the element-wise Hadamard product, 
$\bm{x}\in\mathbb{R^{\mathrm{N\times 1}}}$ is the $1$-dimension signal on graph $\mathcal{G}$ 
and $\mathcal{K}$ is the convolutional kernel on the graph.
The graph Laplacian defined as \mbox{$\bm{L}=\bm{I}_N - \bm{D}^{(-1/2)}\bm{A}\bm{D}^{(-1/2)}$}), where \mbox{$\bm{D}=diag\{d_1,d_2,...,d_{N}\}$} and \mbox{$d_i = \sum_{j=0}^{N} A(i,j)$}.
Equation \eqref{eq:kernel} can be approximated by a truncated expansion with $K$-order Chebyshev polynomials as follows (more details are given in~\cite{HAMMOND2011129}\cite{kipf2016semi}).

\begin{equation}
\label{eq:Chebyshev1}
\begin{split}
\bm{x}\,*_{\mathcal{G}}\,\mathcal{K} \approx & \sum_{k=0}^{K}\theta_kT_k(\hat{\bm{L}})\bm{x} \\
= & \theta_0T_0(\hat{\bm{L}})\bm{x} + \theta_1T_1(\hat{\bm{L}})\bm{x} + \cdots + \theta_KT_K(\hat{\bm{L}})\bm{x},
\end{split}
\end{equation}
where the Chebyshev polynomials are recursively defined by, 
\begin{subequations}
\begin{align}
\label{eq:Chebyshev2}
& T_k(\hat{\bm{L}})=2\hat{\bm{L}}T_{k-1}(\hat{\bm{L}})-T_{k-2}(\hat{\bm{L}}) \\
& \textit{with}\quad T_0(\hat{\bm{L}})=\bm{I}_N,\;T_1(\hat{\bm{L}})=\hat{\bm{L}}, 
\end{align}
\end{subequations}
where $\hat{\bm{L}}={2}\bm{L}/{\lambda_M}-\bm{I}_N$, ${\lambda_M}$ is the the largest eigenvalue of $\bm{L}$,
and $\theta_0,\theta_1,\cdots,\theta_K$ are the Chebyshev coefficients, which parameterize the convolutional kernel $\mathcal{K}$.
Substituting \eqref{eq:Chebyshev2} into \eqref{eq:Chebyshev1}, and considering ${\lambda_M}$ as a learnable parameters,
the truncated expansion can be rewritten as, 
\begin{equation}
\label{eq:polynomials}
\sum_{k=0}^{K}\eta_k\bm{L}^k\bm{x} = \eta_0\bm{I}_N\bm{x} + \eta_1\bm{L}\bm{x} + \cdots + \eta_K\bm{L}^K\bm{x}, 
\end{equation}
where $\eta_0,\eta_1,\cdots,\eta_K$ are learnable parameters.
This polynomials can also be written in a recursive form of
\begin{subequations}
\begin{align}
\label{eq:recursive1}
& \sum_{k=0}^{K}\eta_k\bm{L}^k\bm{x} = \sum_{k=0}^{K}\eta_k^{\prime}T_k^{\prime}(\bm{x}) \\ 
\label{eq:recursive2}
& \textit{with} \quad T_{k}^{\prime}(\bm{x})=\bm{L}\eta_{k-1}^{\prime}T_{k-1}^{\prime}(\bm{x}),\;T_0^{\prime}(x)=x \\
\label{eq:recursive3}
& \textit{and} \quad \eta_k^{\prime}=\eta_k/\eta_{k-1},\;\eta_{0}^{\prime}=\eta_{0}. 
\end{align}
\end{subequations}
and $\eta_0^{\prime},\eta_1^{\prime},\cdots,\eta_K^{\prime}$ are also considered as a series of learnable parameters.
We generalize it to high-dimensional signal $\bm{X}\in\mathbb{R^{\mathrm{N\times d_{I}}}}$ as follows: 
each item in \eqref{eq:recursive1} can be written as $\bm{H}_k=T_{k}^{\prime}(\bm{X})\bm{W}_k$.
Recalling \eqref{eq:recursive2}, the convolutional operation on a graph with a high-dimensional signal can be simplified as $K$-order polynomials and defined as, 
\begin{equation}
\label{eq:highrecursive}
\bm{x}\,*_{\mathcal{G}}\,\mathcal{K} \approx \sum_{k=0}^{K}\bm{H}_k,\;\textit{with}\; \bm{H}_k=\bm{L}\bm{H}_{k-1}\bm{W}_k, \bm{H}_0=\bm{X}\bm{W}_0, 
\end{equation}
where $\bm{H}_k\in\mathbb{R^{\mathrm{N\times d}}}$, and $\bm{W}_k\in\mathbb{R^{\mathrm{d\times d}}}$, which are also learnable and parameterize the convolutional kernel.
Based on the matrix multiplication rule, the $i$-\textit{th} line $\bm{h}_i^{k}$ in $\bm{H}_{k}$ can be written as, 
\begin{equation}
\label{eq:equivalence}
\bm{h}_{i}^k=\sum_{j=1}^N\bm{L}(i,j)(\bm{h}_{j}^{k-1}\bm{W}_{k}). 
\end{equation}
Thus, we can translate convolutional operation into $K$ rounds of propagation and aggregation on graphs.
Specifically, at round $k$, the projection function is $\bm{n}_i^k = \bm{h}_{i}^{k-1}\bm{W}_{k}$, the propagation function is $\bm{m}_{j\rightarrow i}^k = \bm{L}(i,j)\bm{n}_i^k$, the aggregation function is $\bm{h}_i^k=\sum_{j\in N(i)}\bm{m}_{j\rightarrow i}^k$.
In other words, \mbox{$\bm{h}_j^{k-1}$} (or $\bm{h}_i^{k}$) is the $j$-\textit{th} line (or $i$-\textit{th} line) of $\bm{H}_{k}$(or $\bm{H}_{k-1}$) and also the output embedding of $v_i$ (or the input embedding of $v_j$).
Each node propagates its message $\bm{n}_i^k$ to its neighbors, 
and also aggregates the received messages sent from neighbors by summing up the values weighted by the corresponding Laplacian weights.

\subsection{Backwards of GNN}
\label{backwards_gnn}
A general GNN model can be abstracted into $K$ combinations of individual stage and conjunction stage.
Each node on the graph can be treated as a data node and transformed separately in a separated stage, which can be written as, 
\begin{equation}
\label{eq:individual_part}
\bm{n}_{i}^k=f_k(\bm{h}_{i}^{k-1}|\bm{W}_{k}). 
\end{equation}
But the conjunction stage is related to the node itself and its neighbors, without loss of generality, it can be written as
\begin{equation}
\label{eq:neighbor_part}
\bm{h}_{i}^k=g_k(A(i,1)\bm{n}_{1}^k,A(i,2)\bm{n}_{2}^k,\cdots,A(i,N)\bm{n}_{N}^k|\bm{\mu}_{k}), 
\end{equation}
where $A(i,j)$ is the element of the adjacency matrix and is equal to the weight of $e_{i,j}$ (refer to the first paragraph in \ref{GNN}).
The forward formula~\eqref{eq:neighbor_part} can be implemented by message passing, as ${n}_{j}^k$ is propagated from $v_j$ along the edges like $e_{i,j}$ to its neighbor $v_i$.
The final summarized loss is related to the final embedding of all the nodes, so can be written as, 
\begin{equation}
\label{eq:finalloss}
L=l(\bm{h}_{0}^K,\bm{h}_{1}^K,\cdots,\bm{h}_{N}^K). 
\end{equation}
According to the multi-variable chain rule, the derivative of previous embeddings of a certain node is, 
\begin{equation}
\label{eq:derivative_node_embeding1}
\frac{\partial L}{\partial\bm{n}_i^{k}} = \frac{\partial L}{\partial \bm{h}_1^k}a_{1,i}\frac{\partial \bm{h}_1^k}{\partial \bm{n}_i^k} + \frac{\partial L}{\partial \bm{h}_2^k}a_{2,i}\frac{\partial \bm{h}_2^k}{\partial \bm{n}_i^k} + \cdots + \frac{\partial L}{\partial \bm{h}_N^k}a_{N,i}\frac{\partial \bm{h}_N^k}{\partial \bm{n}_i^k}. 
\end{equation}
Thus, \eqref{eq:derivative_node_embeding1} can be rewritten as, 
\begin{equation}
\label{eq:derivative_node_embeding2}
\frac{\partial L}{\partial\bm{n}_i^{k}} = \sum_{j\in N_{in}(i)}a_{j,i}\frac{\partial L}{\partial \bm{h}_j^k}\frac{\partial \bm{h}_j^k}{\partial \bm{n}_i^k}
\end{equation}
So the backward of a conjunction stage also can be calculated by a message passing, 
where each node (\romannumeral1) broadcasts the current gradient ${\partial L}/{\partial \bm{h}_j^k}$ to its neighbors along edges  $e_{j,i}$;
(\romannumeral2) calculates the differential ${\partial \bm{h}_j^k}/{\partial \bm{n}_i^k}$ on each edge $e_{j,i}$, multiplies the received gradient and edge weight $a_{j,i}$, and sends the results (vectors) to the destination node $v_i$;
(\romannumeral3) sums up the received derivative vectors, and obtains the gradient ${\partial L}/{\partial \bm{n}_i^{k}}$.
Meanwhile, the forward and backward of each stage are similar to the normal neural network.
The above derivation can expand to the edge-attributed graph.

\subsection{Derivation in MPGNN}
\label{backwards_tgar}
The previous section gives the derivatives for a general GNN model,
and this part describes the derivation for the MPGNN framework.
As in Algorithm~\ref{alg:cbgnn},
a MPGNN framework contains $K$ passes' procedure of "projection-propagation-aggregation", 
where the projection function (Line~6) can be considered as the implementation of an individual stage, 
while the propagation function (Line~7) and aggregation function for the conjunction stage.
Aggregation function adopts the combination of \eqref{eq:acc} and \eqref{eq:apy}.
If the gradients of the $k$-th layer node embeddings $\partial L/ \partial \bm{h}_{i}^{k}$ are given, the gradients of ($k-1$)-th layer node embeddings and the corresponding parameters are computed as, 

\begin{equation}
\label{eq:gradients_u2}
\frac{\partial L}{\partial \bm{\mu}_k^{(2)}} = \sum_{i=1}^N\frac{\partial L}{\partial \bm{h}_{i}^{k}}\frac{\partial \bm{h}_{i}^{k}}{\partial \bm{\mu}_k^{(2)}}, 
\end{equation}

\begin{equation}
\label{eq:gradients_u1}
\frac{\partial L}{\partial \bm{\mu}_k^{(1)}} = \sum_{i=1}^N\frac{\partial L}{\partial \bm{h}_{i}^{k}}\frac{\partial \bm{h}_{i}^{k}}{\partial \bm{M}_{i}^{k}}\frac{\partial \bm{M}_{i}^{k}}{\partial \bm{\mu}_k^{(1)}}, 
\end{equation}

\begin{equation}
\label{eq:gradients_message}
\frac{\partial L}{{\partial \bm{m}_{j\rightarrow i}^{k}}} = \frac{\partial L}{\partial \bm{h}_{i}^{k}}\frac{\partial \bm{h}_{i}^{k}}{\partial \bm{M}_{i}^{k}}\frac{\partial \bm{M}_{i}^{k}}{\partial \bm{m}_{j\rightarrow i}^{k}}, 
\end{equation}

\begin{equation}
\label{eq:gradients_theta}
\frac{\partial L}{\partial \bm{\theta}_k} = \sum_{i=1}^N\sum_{j\in N_O(i)}\frac{\partial L}{{\partial \bm{m}_{j\rightarrow i}^{k}}}\frac{{\partial \bm{m}_{j\rightarrow i}^{k}}}{\partial \theta_k}, 
\end{equation}

\begin{equation}
\label{eq:gradients_prj}
\begin{split}
\frac{\partial L}{{\partial \bm{n}_{i}^{k}}} = &\sum_{j\in N_O(i)}\frac{\partial L}{{\partial \bm{m}_{j\rightarrow i}^{k}}}\frac{{\partial \bm{m}_{j\rightarrow i}^{k}}}{{\partial \bm{n}_{i}^{k}}} + \\
&\sum_{j\in N_I(i)}\frac{\partial L}{{\partial \bm{m}_{i \rightarrow j}^{k}}}\frac{{\partial \bm{m}_{i\rightarrow j}^{k}}}{{\partial \bm{n}_{i}^{k}}} + \frac{\partial L}{\partial \bm{h}_{i}^{k}}\frac{\partial \textit{Apy}_{k}}{\partial \bm{n}_{i}^{k}}, 
\end{split}
\end{equation}

\begin{equation}
\label{eq:gradients_w}
\frac{\partial L}{\partial \bm{W}_k} = \sum_{i=1}^N\frac{\partial L}{\partial \bm{n}_{i}^{k}}\frac{\partial \bm{n}_{i}^{k}}{\partial \bm{W}_k}, 
\end{equation}

\begin{equation}
\label{eq:gradients_h}
\frac{\partial L}{\partial \bm{h}_i^{k-1}} = \frac{\partial L}{\partial \bm{n}_{i}^{k}}\frac{\partial \bm{n}_{i}^{k}}{\partial \bm{h}_i^{k-1}}.
\end{equation}
\section{Training Strategy Examples}
\label{appendix:batchexample}
\begin{figure*}[t!]
\setlength{\abovecaptionskip}{0cm}
\centering
\subfigure[Nodes \textit{A} and \textit{B} form a batch]{
\begin{minipage}[t]{0.249453\linewidth}
\centering
\includegraphics[width=0.8\linewidth]{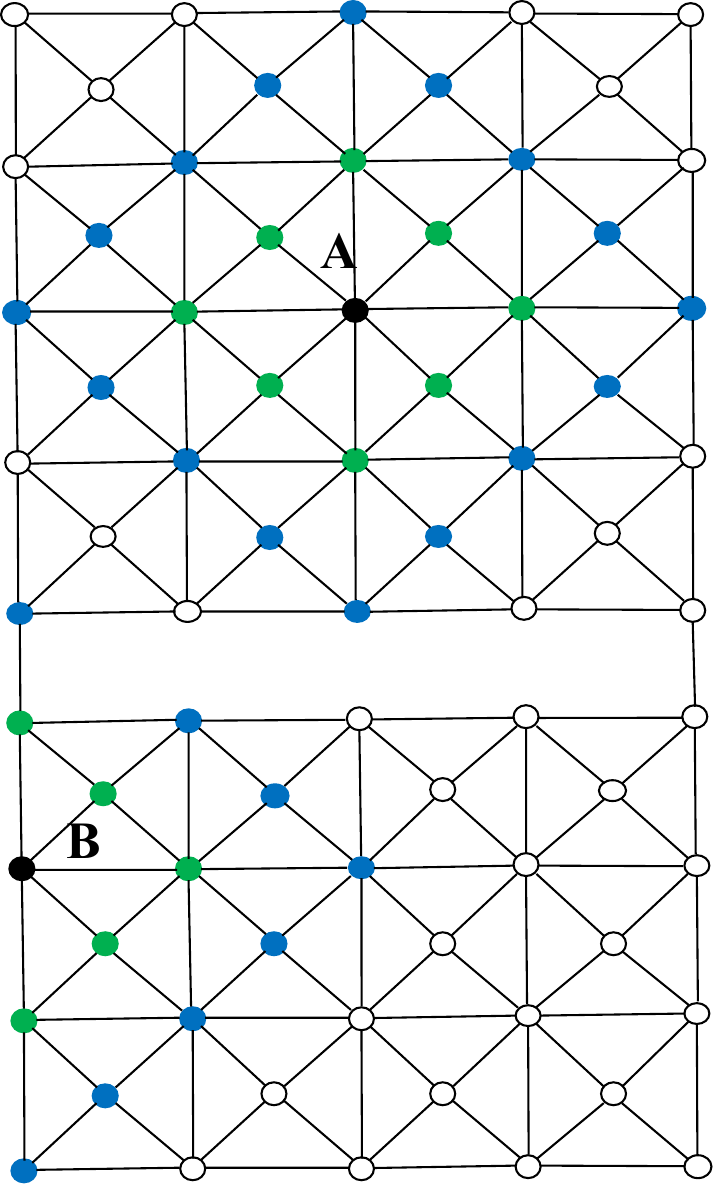}
\label{fig:minibatch1}
\end{minipage}%
}%
\subfigure[Nodes \textit{C} and \textit{D} form a batch]{
\begin{minipage}[t]{0.249453\linewidth}
\centering
\includegraphics[width=0.8\linewidth]{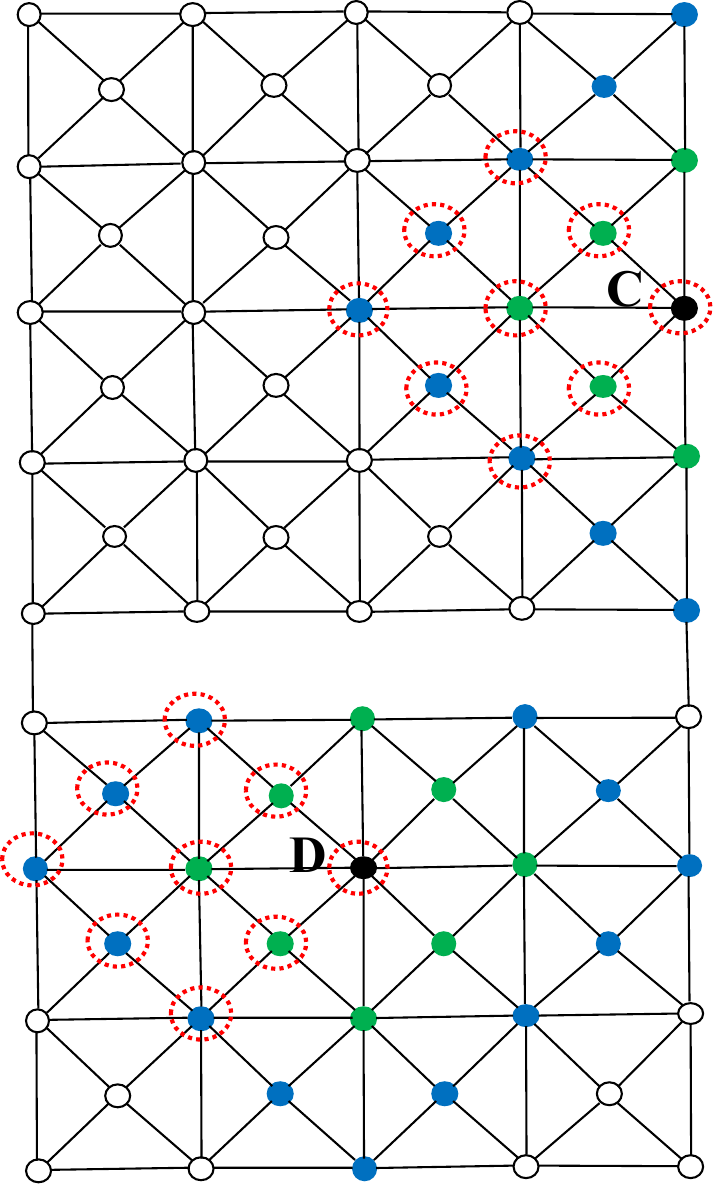}
\label{fig:minibatch2}
\end{minipage}%
}%
\subfigure[Community \textit{A} forms a batch]{
\begin{minipage}[t]{0.25237\linewidth}
\centering
\includegraphics[width=0.8\linewidth]{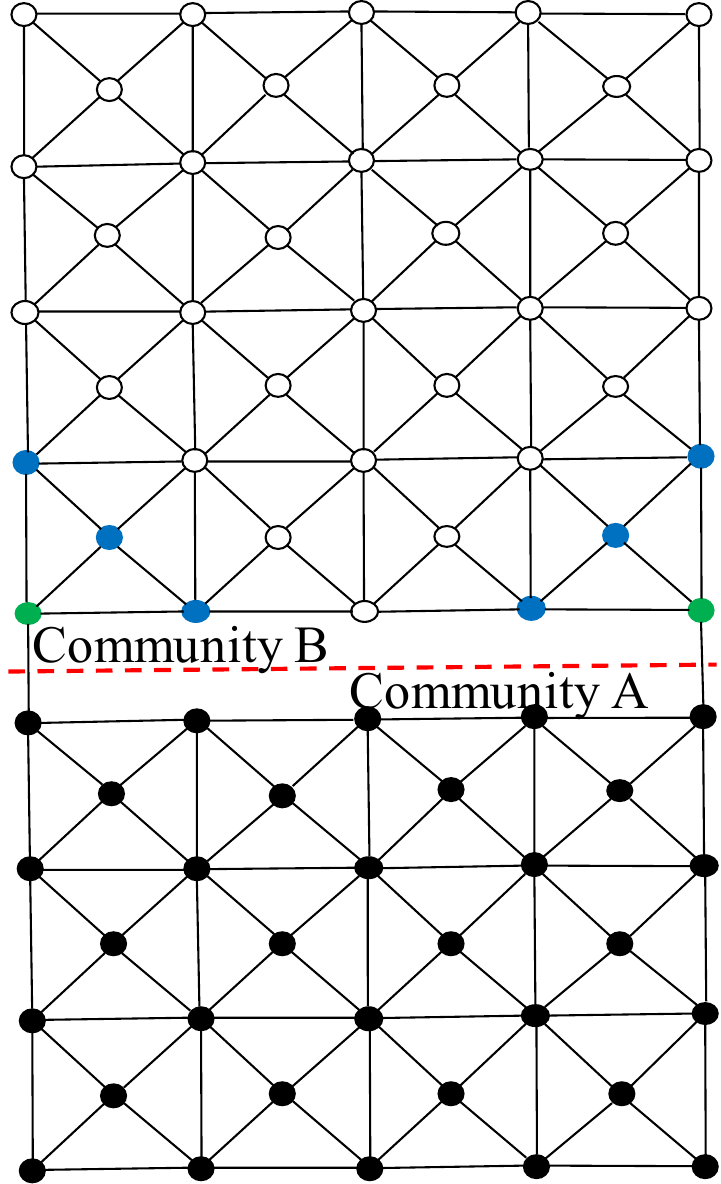}
\label{fig:combatch1}
\end{minipage}
}%
\subfigure[Community \textit{B} forms a batch]{
\begin{minipage}[t]{0.2487\linewidth}
\centering
\includegraphics[width=0.8\linewidth]{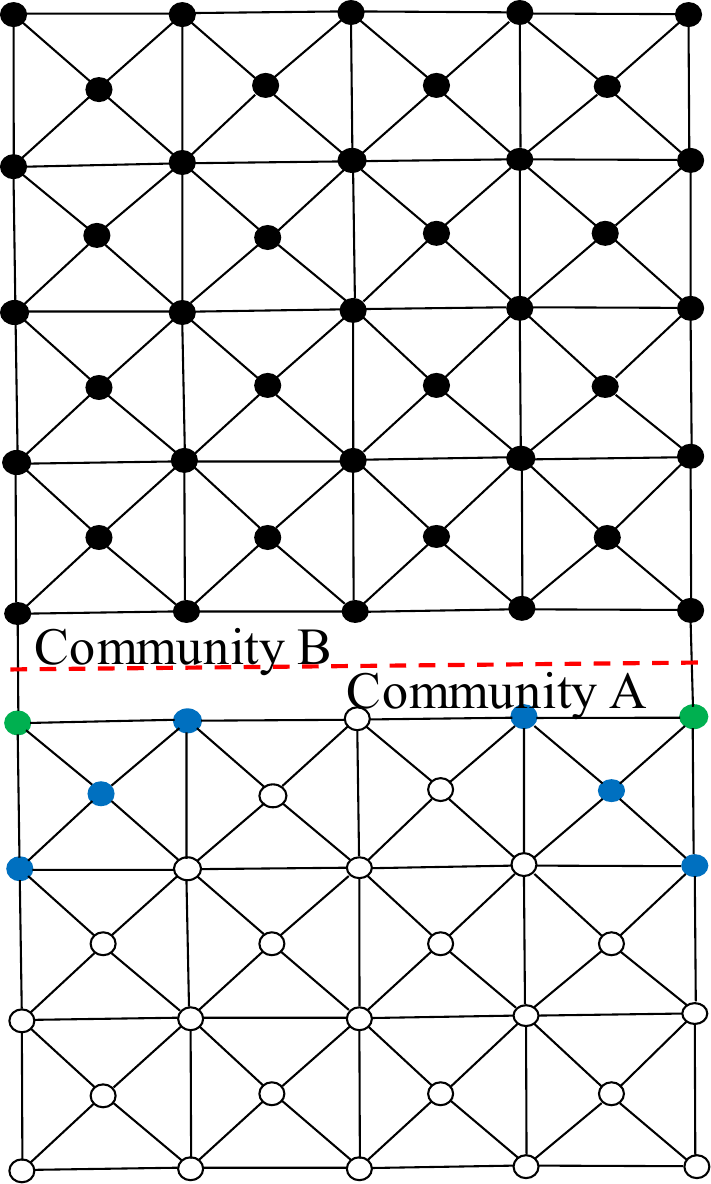}
\label{fig:combatch2}
\end{minipage}
}%
\centering
\caption{An example of a two-hop node classification: (a) and (b) are of mini-batch,  (c) and (d) are of cluster-batch. The black solid points are the target nodes to generate mini-batches. The green and blue solid nodes are the one-hop and two-hop neighbors of their corresponding target nodes. The hollow nodes are ignored during the embedding computation. The nodes with circles are  the shared neighbors between mini-batches.}
\label{fig:batchexample}
\end{figure*}

\begin{table*}[!pt]
\centering
\setlength{\belowcaptionskip}{5pt}%
\caption{Comparison of three GNN training strategies.}
\label{tbl:comparison}
\begin{tabular}{Sl Sl Sl}
\toprule
 Strategies & Advantages & Disadvantages\\
  \midrule
  \begin{tabular}[l]{@{}l@{}} Global-\\batch \end{tabular} &
  \begin{minipage}[l]{0.43\textwidth}
  \begin{itemize}[leftmargin=*,noitemsep]
  \item No redundant calculation;
  \item Stable training convergence.
  \end{itemize}
  \end{minipage} &
  \begin{minipage}[l]{0.45\textwidth}
  \begin{itemize}[leftmargin=*,noitemsep]
  \item The highest cost in one step.
  \end{itemize}
  \end{minipage} \\
  \hline

  \begin{tabular}[l]{@{}l@{}} Mini-\\batch \end{tabular} &
  \begin{minipage}[l]{0.43\textwidth}
  \begin{itemize}[leftmargin=*,noitemsep]
  \item Friendly for parallel computing;
  \item Easy to implement on modern DL frameworks.
  \end{itemize}
  \end{minipage} &
  \begin{minipage}[l]{0.45\textwidth}
  \begin{itemize}[leftmargin=*,noitemsep]
  \item Redundant calculation among batches;
  \item Exponential complexity with depth;
  \item Power law graph challenge.
  \end{itemize}
  \end{minipage} \\
  \hline

  \begin{tabular}[l]{@{}l@{}}Cluster-\\batch\end{tabular}&
  \begin{minipage}[l]{0.43\textwidth}
  \begin{itemize}[leftmargin=*,noitemsep]
  \item Advantages of mini-batch but with less redundant calculation.
  \end{itemize}
  \end{minipage} &
  \begin{minipage}[l]{0.45\textwidth}
  \begin{itemize}[leftmargin=*,noitemsep]
  \item Limited support for graphs without obvious community structures;
  \item Instable learning speed and imbalanced batch size.
  \end{itemize}
  \end{minipage} \\
  \bottomrule
\end{tabular}
\vspace{-0.3cm}
\end{table*}

Figures ~A\ref{fig:combatch1} and A\ref{fig:combatch2} illustrate an example of cluster-batched computation,
where the graph is partitioned into two communities (or clusters).
The black nodes in Figure~A\ref{fig:combatch1} are in community \texttt{A} and the black nodes in Figure~A\ref{fig:combatch2} belong to community \texttt{B}.
The green or blue nodes in both graphs are the 1-hop or 2-hop boundaries of a community.
To train a 2-layered GNN model, we can select community \texttt{A} as the first batch and community \texttt{B} as the second.
In this case, to achieve more flexibility, our system allows users to configure whether to get their 1-hop or 2-hop boundary neighbors involved in the embedding computation of the black nodes in communities \texttt{A} and \texttt{B}.
By default, this feature is disabled, i.e. only nodes in the communities participate in the computation as done in Cluster-GCN.

We compare the pros and cons of global-batched, mini-batched, and cluster-batched training strategies in Table~\ref{tbl:comparison}.
These comparisons imply the necessity to design a new GNN learning system that enables the exploration of different training strategies and a solution to address the limitations of existing architectures.
%
\section{Accuracy Comparison with GAT Model}
\label{appendix:gat}
\begin{table}[t!]
\centering
\caption{Accuracy comparison with GAT model.}
\label{appendix:tab_gat}
\begin{tabular}{cccc}
\toprule
                                      & \multicolumn{3}{c}{Accuracy in Test Set (\%)}                    \\ \cline{2-4} 
\multirow{-2}{*}{Dataset} & \begin{tabular}[l]{@{}r@{}} \projectname w/GB \end{tabular} 
                          & \begin{tabular}[l]{@{}r@{}} \projectname w/MB \end{tabular}  
                          & \begin{tabular}[l]{@{}c@{}} DGL \end{tabular}    \\ 
\midrule                           
Cora         &  81.1 & 80.0 & \textbf{81.4}	\\ 
Citeseer     &  71.2 & 70.8& \textbf{72.6} \\ 
Pubmed       &  \textbf{78.7} & 78.6 & 78.0	\\
\bottomrule
\multicolumn{4}{l}{GB: global-batch, MB: mini-batch.}
\end{tabular} 
\end{table}
In the main text, we have used the popular GCN algorithm for performance comparison between our system and existing DL frameworks.
As stated in the main text, the purpose of these tests is to show that our system can learn GNNs as well as existing frameworks.
Herein, we use the GAT model as another example algorithm and three publicly available datasets, i.e. Cora. Citeseer and Pubmed, to compare the performance between our system and DGL for readers' information.
Table~\ref{appendix:tab_gat} shows the accuracy comparison with the GAT model, where it can be observed that our system yields comparable accuracy with DGL.
\section{Performance Assessment}
\subsection{Scalability Assessment}
\label{appendix:sec_distgdl_scale}
In this section, we show the detailed scalability of DistDGL~\cite{zheng2020distdgl, zheng2022distributed} by means of training four GCN models of a different number of layers on the Reddit dataset.
Same with the tests in the main text, the number of layers ranges from 2 to 5, and node sampling is disabled for a fair comparison.
Table~\ref{tbl:distgdl_scalability} shows the runtime per mini-batch in the function of the number of trainers.
From the table, the runtime increases as the number of trainers grows, indicating that DistDGL does not scale at all concerning number of trainers.
Further, except for the 2-layered model, all other models encountered socket errors when the number of trainers is large.
Specifically, the 3-layered model failed in 128 trainers, while both the 4-layered and 5-layered models started to fail from 64 trainers.
\begin{table}[t!]
\centering
\caption{Runtimes (in seconds) of DistDGL in the function of the number of trainers.}
\label{tbl:distgdl_scalability}
\begin{tabular}{cccccc}
\toprule
\multirow{2}{*}{\#Trainers} & \multicolumn{4}{c}{Number of layers}\\
\cline{2-5}
& 2 & 3 & 4 & 5 \\
\midrule
8  &22.031	&55.775	&88.941	&123.507    \\
16  &22.793	&60.959	&99.537	&137.468    \\  
32	&25.458	&73.233	&124.891	&174.956    \\
64	&30.967	&105.259	&Socket Error	&Socket Error   \\
128	&41.301	&Socket Error	&Socket Error	&Socket Error   \\
\bottomrule
\end{tabular}
\end{table}

We would like to express that we have excluded Dorylus~\cite{thorpe2021dorylus} and ROC~\cite{jia2020improving} from this comparison for the following reasons. 
Firstly, Dorylus is based on AWS Lambda and cannot run in our Kubernetes cluster.
Secondly, Dorylus performs graph operations in distributed graph servers and conducts neural network operations by Lambda threads. However, it still needs to load all neighbors of given target nodes into Lambda threads. This is the same as DistDGL.
Thirdly, Dorylus runs slower than DGL (non-sampling) on Reddit as shown in its paper.
Finally, ROC requires each graph server to load the entire graph into memory and thus does not match the objective of our tests for fully distributed graphs.

\subsection{Parameter Tuning}
\label{appendix:sec_distdgl_tune}
\begin{figure}[t!]
\centering
\subfigure[2-layered GCN]{
\begin{minipage}[c]{0.45\linewidth}
\includegraphics[width=\linewidth]{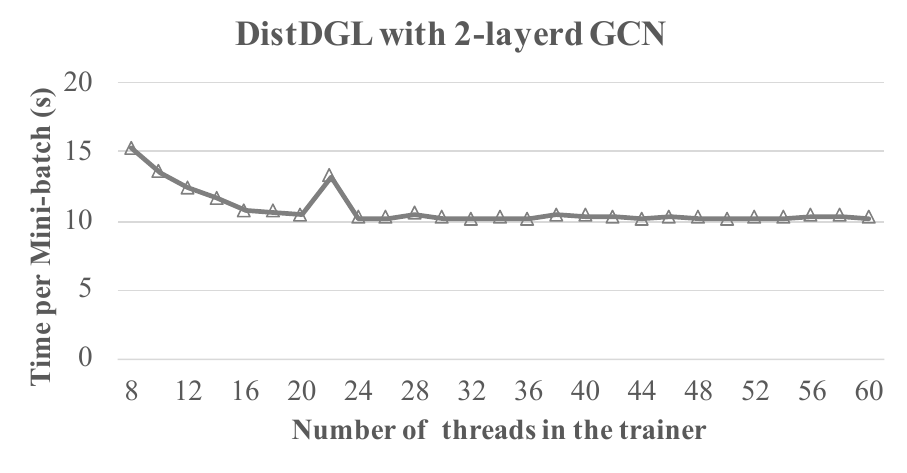}
\label{fig:distdgl_2layered}
\end{minipage}
}
\subfigure[3-layered GCN]{
\begin{minipage}[c]{0.45\linewidth}
\includegraphics[width=\linewidth]{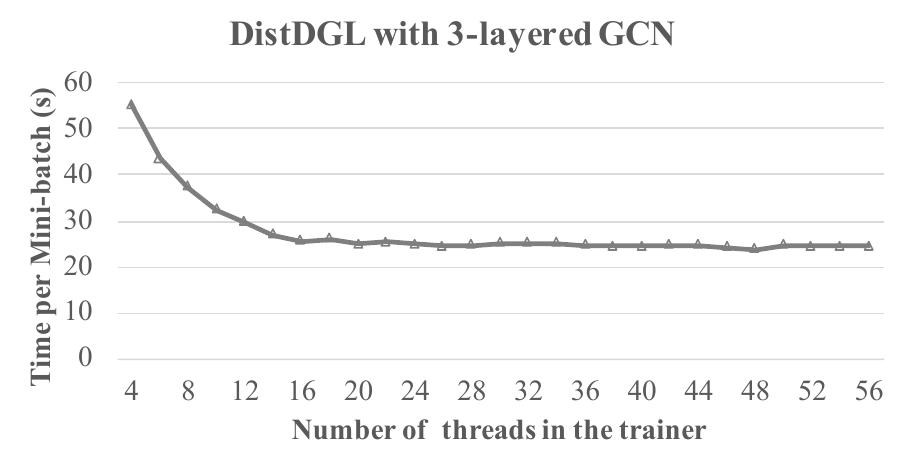}
\label{fig:distdgl_3layered}
\end{minipage}
}

\subfigure[4-layered GCN]{
\begin{minipage}[c]{0.45\linewidth}
\includegraphics[width=\linewidth]{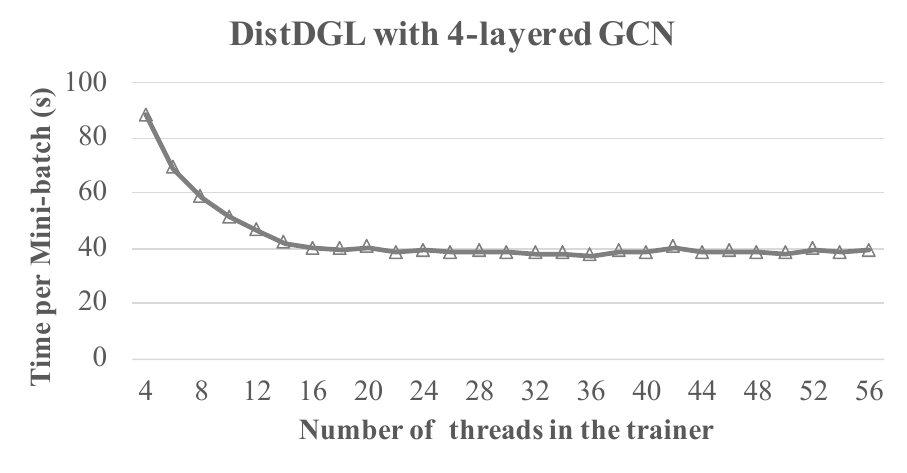}
\label{fig:distdgl_4layered}
\end{minipage}
}
\subfigure[5-layered GCN]{
\begin{minipage}[c]{0.45\linewidth}
\includegraphics[width=\linewidth]{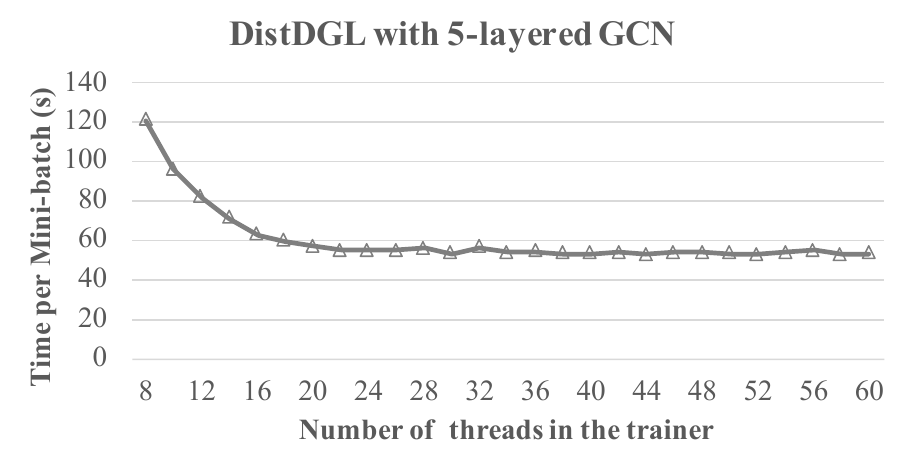}
\label{fig:distdgl_5layered}
\end{minipage}
}
\caption{Parameter tuning of DistDGL for 4 GCN models on Reddit.}
\label{fig:distdgl_tuning}
\end{figure}

As mentioned in the main text, we launch only one trainer in each machine and tune the number of threads assigned to the trainer and the distributed server.
In this test, in each machine, the number of threads assigned to the server is calculated as $64-p$, where $p$ denotes the number of threads used by the trainer. Figure~\ref{fig:distdgl_tuning} illustrates the performance changes along with the tuning of value $p$ for each GCN model.
According to our tuning, the best performance of DistDGL is gained by setting $p=44$ for the 2-layered model, $p=48$ for the 3-layered model, $p=36$ for the 4-layered model, and $p=58$ for the 5-layered model.
\subsection{Ablation Study}
\label{appendix:ablation_study}
\begin{figure}[!t]
\begin{minipage}{\linewidth}
\centering
\includegraphics[width=\linewidth]{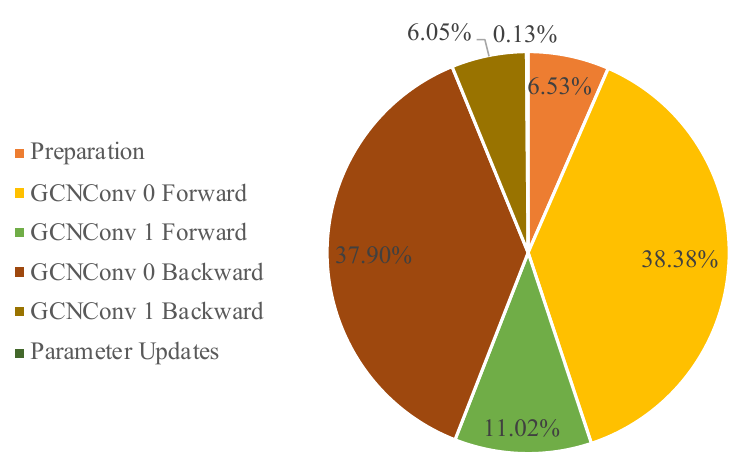}
\caption{Ablation study of \projectname on the runtime percentage of each stage of 2-layered GCN on ogbn-papers100M.}
\label{fig:graphtheta_abalation}
\end{minipage}
\vspace{-0.2cm}
\end{figure}

We have conducted an ablation study of \projectname on the runtime percentage of each stage of the 2-layered GCN on the ogbn-papers100M dataset.
In this test, we use the mini-batched training strategy and launch 128 workers in our cluster.
The training procedure of a mini-batch step is split into six phases for break-down analysis, i.e. preparation, the forward computation of graph convolution (GCNConv) layer 0, the forward computation of GCNConv layer 1, the backward computation of  GCNConv layer 0, the backward computation of GCNConv layer 1, and parameter updates for the whole model.
Figure~\ref{fig:graphtheta_abalation} illustrates the runtime percentage of each phase.
From the figure, we can see that the forward and backward computation of GCNConv layer 0 dominates the overall runtime with a total percentage of up to 76.28\%.
This is consistent with our expectation, as this layer processes the most nodes and edges.
\end{appendices}

\end{document}